\documentclass[10pt,journal,compsoc]{IEEEtran}
\usepackage{amsmath,amssymb,bbold}

\usepackage{bauer, bauer_e}
\usepackage{algorithm,algorithmic}
\usepackage{hyperref}
\usepackage{breqn}
\usepackage{booktabs}

\newenvironment{proof}[1][Proof:\\]{\begin{trivlist}
\item[\hskip \labelsep {\bfseries #1}]}{\end{trivlist}$\Box$}

\usepackage{color, colortbl}
\newcommand{\blueOn}{\color{blue}}

\newcommand{\redOn}{\color{red}}
\newcommand{\brownOn}{\color{brown}}
\newcommand{\blackOn}{\color{black}}
\definecolor{LightCyan}{rgb}{0.88,1,1}
\definecolor{ColumnColor}{rgb}{1.0,0.941454,0.955287}
\definecolor{brown}{rgb}{0.58823, 0.294117, 0}

\graphicspath{{./}}

\ifCLASSOPTIONcompsoc
  \usepackage{cite} 	 		  
\else
  \usepackage{cite}
\fi
\ifCLASSINFOpdf
\else
\fi
\usepackage{amsmath}
\begin{document}
%
\title{Self-Supervised Training with Autoencoders\\ for Visual Anomaly Detection}

%
%
%
%

\author{Alexander~Bauer, Shinichi~Nakajima, Klaus-Robert~M{\"u}ller
\thanks{
A. Bauer is with BASLEARN -- TU Berlin/BASF Joint Lab for Machine Learning, Technische Universität Berlin, 10587 Berlin, Germany; alexander.bauer@tu-berlin.de.\\
S. Nakajima is with Berlin Institute for the Foundations of Learning and Data, Technische Universität Berlin, 10587 Berlin, Germany,
and with RIKEN Center for AIP, Japan; nakajima@tu-berlin.de.\\
K.-R. M{\"u}ller is with Berlin Institute for the Foundations of Learning and Data, Technische Universität Berlin, 10587 Berlin, Germany,
and with Department of Artificial Intelligence, Korea University, Anam-dong, Seongbuk-gu, Seoul 02841, Korea,
and with Max-Planck-Institut für Informatik, 66123 Saarbrücken, Germany; klaus-robert.mueller@tu-berlin.de.
}
\thanks{(Corresponding authors: Alexander Bauer, Klaus-Robert M{\"u}ller)}
}

\IEEEtitleabstractindextext{%
\begin{abstract}
We focus on a specific use case in anomaly detection where the distribution of normal samples is supported by a lower-dimensional manifold.
Here, regularized autoencoders provide a popular approach by learning the identity mapping on the set of normal examples,
while trying to prevent good reconstruction on points outside of the manifold.
Typically, this goal is implemented by controlling the capacity of the model,
either directly by reducing the size of the bottleneck layer or implicitly by imposing some sparsity (or contraction) constraints on parts of the corresponding network.
However, neither of these techniques does explicitly penalize the reconstruction of anomalous signals often resulting in poor detection.
We tackle this problem by adapting a self-supervised learning regime that exploits discriminative information during training but focuses on the submanifold of normal examples.
Informally, our training objective regularizes the model to produce locally consistent reconstructions,
while replacing irregularities by acting as a filter that removes anomalous patterns.
To support this intuition, we perform a rigorous formal analysis of the proposed method
and provide a number of interesting insights.
In particular, we show that the resulting model resembles a non-linear orthogonal projection
of partially corrupted images onto the submanifold of uncorrupted samples.
On the other hand, we identify the orthogonal projection
as an optimal solution for a number of regularized autoencoders
including the contractive and denoising variants.
Among the idempotent mappings, orthogonal projection further provides a certain conservation effect on its arguments
by projecting the anomalous samples in a way largely preserving the original content.
This, in turn, enables accurate detection and localization of the anomalous regions by means of the reconstruction error.
We support our theoretical analysis by empirical evaluation of the resulting detection and localization performance of the proposed method.
In particular, we achieve a new state-of-the-art result on the MVTec AD dataset -- a challenging benchmark for visual anomaly detection in the manufacturing domain.
\end{abstract}

\begin{IEEEkeywords}
Autoencoders, anomaly detection, manifold learning.
\end{IEEEkeywords}}

\maketitle

\IEEEdisplaynontitleabstractindextext

%
\IEEEpeerreviewmaketitle

\section{Introduction}
\label{sec:section1}
\IEEEPARstart{T}{he} task of anomaly detection (AD) in a broad sense corresponds to searching for patterns,
which considerably deviate from some concept of normality.
The criteria for what is normal and what is an anomaly can be very subtle and depends heavily on the application.
Visual AD specifically aims to detect and locate anomalous regions in imagery data
with practical applications in the industrial,
medical and other domains \cite{HaselmannGT18, BergmannLFSS19, LSR, BergmannFSS20, VenkataramananP20, SchleglSWSL17, NapoletanoPS18, texture2016, LiuLZKWBRC20, RothPZSBG22, WanGLW22, abs-1806-04972, SchleglSWLS19, TanHDSRK21, ZimmererIPKM19, AbatiPCC19, DefardSLA20}.
The continuous research in this area produced a variety of methods
ranging from classical unsupervised approaches like PCA  \cite{Hotelling33, ScholkopfSM98, Hoffmann07}, One-Class SVM \cite{ScholkopfPSSW01},  SVDD \cite{TaxD04},
nearest neighbor algorithms \cite{KnorrNT00, RamaswamyRS00}, and KDE \cite{Parzen62}
to more recent methods including
various types of  autoencoders \cite{PrincipiVSP17, abs-1806-04972, ChalapathyMC17, KieuYGJ19, ZhouP17, ZongSMCLCC18, KimSLJCKY20, DengZMS13, VenkataramananP20},
deep one-class classification \cite{ErfaniRKL16, KimKYC23, RuffGDSVBMK18},
generative models \cite{SchleglSWSL17, SchleglSWLS19},
self-supervised approaches \cite{GolanE18, TackMJS20, LSR, HaselmannGT18, ZavrtanikKS212, ZavrtanikKS21, LiSYP21, InTra}
and others \cite{NapoletanoPS18, texture2016, LiuLZKWBRC20, LeeLS22, abs-2211-07381, abs-2211-12634, TsaiWL22, ZouJPZD22, LiJMWHG21, WanGLW22, SalehiSBRR21, DengL22, RudolphWRW23, CaoWSG22, ZhangWK22, WanGLW23, WanCGSL22, ZhengWDBZW22, GudovskiyIK22, abs-2210-14913, YiY20, HuCS21, YangWF23, YanWZC21, CollinV20, TaoZMHLA22, abs-2210-14485, KimJKCKC22, abs-2205-06568, LiznerskiRVFKM21, LSR, LiSYP21, InTra, RothPZSBG22, Jaehyeok2023, SchluterTHK22, DehaeneFCE20, abs-2110-03396, LeeK22c, JiangZBCKDSX23, WuCFL21, JiangLWNWLWZ22, RuffKVMSKDM21}.
\begin{figure}
\centering
\includegraphics[scale = 0.465]{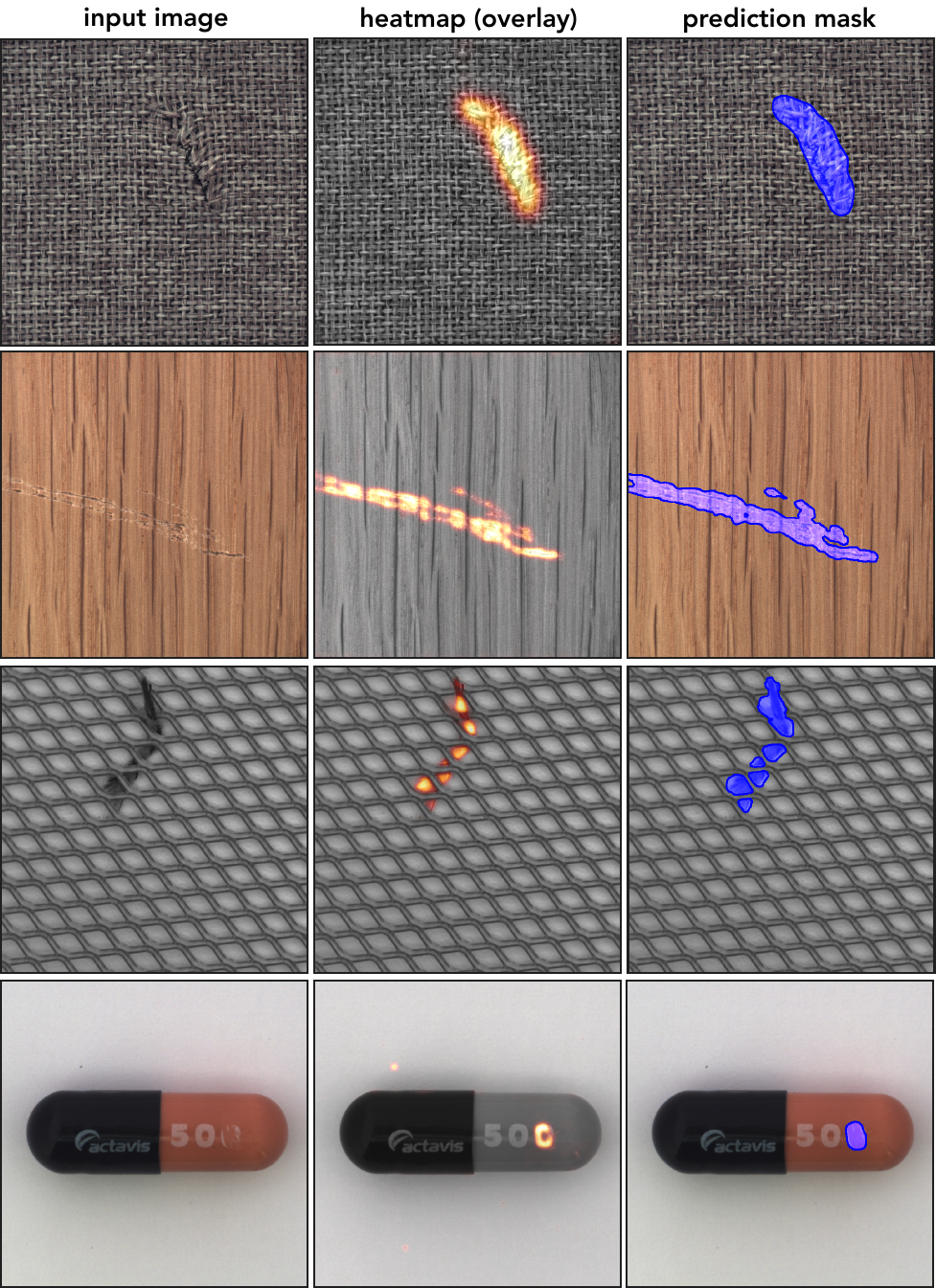}
\caption{
A few anomaly detection results of our approach.
Each row shows the input image, an overlay with the anomaly heatmap, and the resulting prediction mask, respectively.
}
\label{fig_bauer1}
\end{figure}

The one-class classifiers \cite{ScholkopfPSSW01, TaxD04}, for example, 
aim at learning a tight decision boundary around the normal examples in the feature space
and define a distance-based anomaly score relative to the center of the training data.
The success of this approach strongly depends on the availability of suitable features.
Therefore, it usually allows for detection of outliers only, which greatly deviate from the normal structure.
In practice, however, we are often interested in more subtle deviations,
which require a good representation of the data manifold.
The same applies to the combined approach based on training deep neural networks (DNNs) directly
with the one-class objective \cite{RuffGDSVBMK18, ChongRKB20, HuFKTO21}.
Although this objective encourages  the network to concentrate the training data in a small region in the feature space,
there is no explicit motivation for anomalous examples to be mapped outside of the decision boundary.
In other words, the one-class objective  gives preference to models, which map its input domain to a narrower region
and does not explicitly focus on separating anomalous examples from the normal data.

Recently, deep autoencoders (AEs) have been used for the task of anomaly detection in the visual domain \cite{HaselmannGT18, LiSYP21, InTra}.
Unlike one-class based approaches, they additionally provide a way for locating the anomalous regions in images
by leveraging the pixel-wise nature of a corresponding training objective.
By optimizing for the reconstruction error using anomaly-free examples, the common belief is
that a corresponding network should fail to accurately reconstruct anomalous regions in the application phase.
Typically, this goal is implemented by controlling the capacity of the model, either directly by reducing the size of the bottleneck layer
or implicitly by imposing some sparsity or contraction constraints on parts of the corresponding network \cite{NIPS2007_c60d060b, LeNCLPN11, RifaiVMGB11, AlainB14, DengZMS13}.
However, neither of these techniques does explicitly penalize reconstruction
of anomalous signals often resulting in a poor detection.
This is similar to the issue related to training with the one-class objective,
where no explicit mechanism exists for preventing anomalous examples from being mapped to the normal region.
In fact, an unsupervised trained AE aims to compress and accurately reconstruct
the input images and does not care much about the actual distinction between normal and anomalous samples.
As a result, the reconstruction error for the anomalous and normal regions can be very similar preventing reliable detection and localization.

In this paper, we propose a self-supervised learning framework,
which introduces discriminative information during training to prevent
good reconstruction of anomalous patterns in the testing phase.
We begin with an observation that the abnormality of a region in the image can be implicitly characterized
by how well it can be reconstructed from the context of the surrounding pixels.
Imagine an image where a small patch has been cut out by setting the corresponding pixel values to zeros.
We can try to reconstruct this patch by interpolating the surrounding pixel values
according to our knowledge about the patch distribution of the training data.
If the reconstruction significantly deviates from the original image,
we consider a corresponding patch to be anomalous and normal otherwise.
Following this intuition, we feed partially distorted images to the network during training
while forcing it to recreate the original content --
similar to the task of neural image completion \cite{PathakKDDE16, IizukaS017, Yu0YSLH18, LiuRSWTC18, YuLYSLH19, abs-1802-05798}.
However, instead of setting the individual pixel values to zeros -- as for the completion task -- we apply a patch transformation,
which approximately preserves the color distribution of the distorted regions, making it more difficult for the network to realize which image regions have been modified.
In order to succeed, the network now must accomplish two different tasks:
a) detection of image regions deviating from the expected pattern and
b) recreation of  the original content from the neighboring pixel sets.
Altogether, the resulting training effect regularizes the model to produce locally consistent reconstructions while replacing irregularities,
therefore, acting as a filter that removes anomalous patterns.
Figure \ref{fig_bauer1} illustrates a few examples to  
give a sense of visual quality of the resulting prediction masks when training according to our approach.

To support our intuition, we perform a theoretical analysis of the proposed method and provide a number of interesting insights.
In particular, we show that with the increasing number of image pixels, the resulting model converges
to the (non-linear) orthogonal projection of partially corrupted images onto the submanifold of normal samples.
Given the orthogonal projection, we can naturally measure the abnormality of anomalous samples
according to their distance from the submanifold.
Additionally, we investigate the effect of projection mappings on the segmentation accuracy of anomalous regions.
Specifically, we analyze the conservation property of the orthogonal projection,
which removes anomalous patterns in a way largely preserving the original content.
As our final result, we establish a close connection of our approach to the previous autoencoding models
based on the assumption that the target distribution concentrates in the vicinity of a lower-dimensional manifold embedded in the input space.
More precisely, we show that the orthogonal projection provides an optimal solution for a number of regularized AEs including the contractive and denoising variants.

The rest of the paper is organized as follows.
In Section \ref{sec:section2}, we discuss the main differences of our approach to the related works
and summarize the main contributions.
In Section \ref{sec:section3}, we formally introduce the proposed framework including our training objective, data generation, and model architecture
followed by a deeper theoretical analysis in Section \ref{sec:section4}.
In Section \ref{sec:section5}, we evaluate the performance of our approach and provide a conclusion in Section \ref{sec:section6}.

\section{Related Works}
\label{sec:section2}
The plethora of existing AD methods (see \cite{RuffKVMSKDM21} for an overview) can be roughly divided in three main groups:
methods based on probabilistic models, one-class classification methods, and methods based on reconstruction models,
while our approach is a member of the latter.
Without going deeper into the details we note that
the first two groups appear less suitable in our use case, where the data distribution is supported by a lower-dimensional manifold.
Due to this fact the Lebesgue measure of the data manifold is zero
excluding the existence of a density function over the input space, which only takes on positive values on the manifold.
On the other hand, methods based on the one-class objective implicitly assume a non-zero volume of the corresponding data
in order to provide accurate detection results,
which again violates our assumption on the data distribution.

Here, we focus on the reconstruction methods represented by the regularized AEs including the contractive and denoising variants
and a number of self-supervised methods inspired by the task of neural image completion.
The central idea behind the contractive autoencoder (CAE), for example, is to learn a reconstruction mapping $r(\bfx) = d(e(\bfx))$
composed from encoder and decoder part
by adding a regularization term $\|\partial_{\bfv} e(\bfx)\|_2$ to the objective,
which implicitly promotes a reduction in magnitude of gradients in direction $\bfv$
pointing outside the tangent plane on a corresponding data manifold at point $\bfx$.
In contrast, the denoising autoencoder (DAE) aims to minimize the reconstruction error without regularization terms,
but augments the training procedure by adding a stochastic noise to the inputs.
The goal here is to learn a model, which can reconstruct the original inputs from noisy samples.
Previous work \cite{AlainB14} has demonstrated that 
the DAE with small noise corruption of variance $\sigma^2$ is similar to the CAE with penalty coefficient $\lambda = \sigma^2$
but where the contraction is imposed explicitly on the whole reconstruction function rather than on the encoder part alone.
Our approach also aims to train a model to reconstruct original content from modified inputs
but differs from the  DAE by more involved input modifications beyond the i.i.d. corruption noise.
Therefore, it can be considered as a generalization of the DAE (and therefore also of the CAE)
to arbitrary complex input modifications.
In particular, the corruptions introduced during training
can be either fully, partially, or non-stochastic at all allowing for a wider range of anomalous patterns,
which can be detected during the application phase.

Another group of reconstruction-based methods, which are related to our approach,
has been inspired by the task of neural image completion \cite{PathakKDDE16, IizukaS017, Yu0YSLH18, LiuRSWTC18, YuLYSLH19, abs-1802-05798}.
This includes a number of self-supervised methods like
LSR  \cite{LSR}, RIAD \cite{ZavrtanikKS21}, CutPaste  \cite{LiSYP21}, InTra  \cite{InTra}, DRAEM \cite{ZavrtanikKS212}, SimpleNet \cite{LiuZXW23},
which (similar to our approach) aim at training a model
on corrupted inputs to reconstruct the original content either in the latent or the input space.
The main differences of these methods to our approach are in the choice of data augmentation during training
and specific network architecture of our model,
which in summary lead to a higher performance in our experiments on the MVTec AD dataset \cite{BergmannBFSS21}.
Another two recent works (\emph{PatchCore}) \cite{RothPZSBG22} and (\emph{PNI}) \cite{Jaehyeok2023} report impressive results on this benchmark.
Both are based on usage of memory banks comprising locally aware nominal patch-level feature representations extracted from pre-trained networks.
In contrast, our approach neither requires pre-training nor additional storage place for the memory bank and performs on-par with the two methods.
In the following we briefly summarize our contributions.

As our \textbf{first contribution},
we formulate an effective AD framework
for the special (and frequent) case,
where the normal examples live on a lower-dimensional submanifold embedded in the input space.
While similar ideas have been used across different domains,
our approach distinguishes itself by the specific form of data augmentation and the network architecture
that together produce a model,
which consistently outperforms previous methods.
In particular, we achieve a new state-of-the-art result for both detection and localization tasks
on the MVTec AD dataset -- a challenging benchmark for visual anomaly detection in the manufacturing domain.

As our \textbf{second contribution},
we provide a rigorous theoretical analysis of the proposed approach
and show that the corresponding model resembles the orthogonal projection
of partially corrupted images onto the submanifold of uncorrupted (anomaly-free) examples.
This covers various forms of input modifications including partial replacements and additive noise.
On the other hand, we show that the orthogonal projection
acts as a filter for irregularities by removing anomalous patterns from the inputs.
Therefore, we conclude that the orthogonality of a projecting model provides (loosely speaking)
a necessary and sufficient condition for accurate anomaly detection and segmentation.

As our \textbf{third contribution}, we establish close connections of our approach to the previous autoencoding models.
Precisely, we identify the (non-linear) orthogonal projection as an optimal solution
minimizing the training objective of the reconstruction contractive autoencoder (RCAE), which is closely related to the CAE and DAE.
This provides a unified view on the group of regularized AEs
based on the assumption that the target distribution concentrates in the vicinity of a lower-dimensional manifold embedded in the input space.

\section{Methodology}
\label{sec:section3}
In the following, we formally introduce our self-supervised framework for training a model to detect and localize anomalous regions in images.
More precisely, we describe the objective function, the form of input modifications during training and our choice of the model architecture.

\subsection{Training Objective}
We identify the map $f \colon [0, 1]^{h \times w \times 3} \rightarrow [0, 1]^{h \times w \times 3}$, $h,w \in \mathbb{N}$
with a DNN that implements an autoencoder with input and output tensors corresponding to color images.
Furthermore, we denote by $\bfx \in [0, 1]^{h \times w \times 3}$ an original (anomaly-free) image
and by $\hat{\bfx}$ a copy of $\bfx$, which has been partially modified.
The modified regions within $\hat{\bfx}$ are encoded through a real-valued mask $\bfM \in [0, 1]^{h \times w \times 3}$,
while $\bar{\bfM} := \mathbf{1} - \bfM$ denotes a corresponding complement with $\mathbf{1} \in \{1\}^{h \times w \times 3}$ being a tensor of one's.
Technically, we train a parameterized model $f_{\theta}$ to project arbitrary inputs from the data space to the submanifold of normal images
by minimizing the following objective
\begin{equation}
\label{E_12041904}
\begin{aligned}
\mathcal{L}(\hat{\bfx}, \bfx, \bfM) =  \frac{(1 - \lambda)}{\|\bar{\bfM}\|_1} \cdot \|\bar{\bfM} \odot \left(f_{\bftheta}(\hat{\bfx}) - \bfx\right) \|_2^2 \hspace*{5pt}+ \\
\frac{\lambda}{\|\bfM\|_1} \cdot \|\bfM \odot \left(f_{\bftheta}(\hat{\bfx}) - \bfx\right)\|_2^2,
\end{aligned}
\end{equation}
where $\odot$ denotes an element-wise tensor multiplication
and $\lambda \in [0, 1]$ is a  hyperparameter controlling the importance of the two terms during training.
Here, $\|\cdot\|_p$ denotes the $l^p$-norm on a corresponding tensor space.
In terms of supervised learning, we feed a partially corrupted image $\hat{\bfx}$ as input to the network,
which is trained to recreate the original image $\bfx$ representing the ground truth label.

\begin{figure}[b]
\centering
\includegraphics[scale = 0.7]{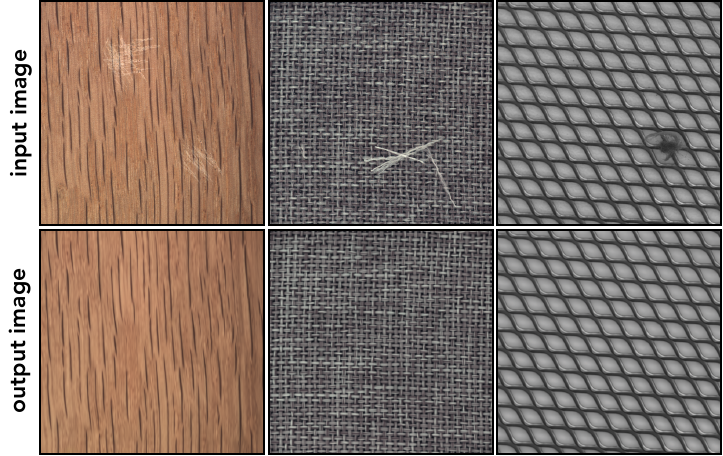}
\caption{
Illustration of the reconstruction effect of our model trained either on the wood, carpet or grid images
(without defects) from the MVTec AD dataset.
}
\label{fig_bauer2}
\end{figure}

By minimizing the above objective, a corresponding autoencoder aims to interpolate between two different tasks.
The first term steers the model to reproduce uncorrupted image regions $\bar{\bfM} \odot \bfx$,
while the second term requires the network to correct the corrupted regions $\bfM \odot \hat{\bfx}$ by recreating the original content.
Altogether, the objective in (\ref{E_12041904}) encourages the model
to produce locally consistent reconstruction of the input while replacing irregularities -- acting as a filter for anomalous patterns.
Figure \ref{fig_bauer2} shows a few reconstruction examples produced by our model $f_{\bftheta}$.
We can see how normal regions are accurately replicated,
while irregularities (e.g., scratches or threads) are replaced by locally consistent patterns.
During training, we use a specific procedure to generate corrupted images $\hat{\bfx}$ from normal examples $\bfx$
based on randomly generated masks $\bfM$ marking the corrupted regions.
We provide a detailed description of this process in the next section.

Given a trained model $f_{\bftheta}$, we can perform AD on an input image $\hat{\bfx}$ as follows.
First we compute the difference map $\text{Diff}[\hat{\bfx}, f_{\bftheta}(\hat{\bfx})]  \in \mathbb{R}^{h \times w}$
between the input $\hat{\bfx}$ and its reconstruction $f_{\bftheta}(\hat{\bfx})$
by averaging the pixel-wise squared difference $(\hat{\bfx} - f_{\bftheta}(\hat{\bfx}))^2$ over the color channels according to:
\begin{equation}
\label{eq_diffmap}
\text{Diff}[\hat{\bfx}, f_{\bftheta}(\hat{\bfx})] := \frac{1}{3}\sum_{c=1}^3 \left[(\hat{\bfx}_{*,*,c} - f(\hat{\bfx})_{*,*,c})^2 \right].
\end{equation}
The binary segmentation mask can be computed by thresholding the difference map.
To get a more robust result we smooth the difference map before thresholding
according to the following formula:
\begin{equation}
\label{eq_anomap}
\text{anomap}_{f_{\bftheta}}^{n,k}(\hat{\bfx}) := G_k^n(\text{Diff}[\hat{\bfx}, f_{\bftheta}(\hat{\bfx})]),
\end{equation}
where $G_k^n = \underbrace{G_k \circ \cdots \circ G_k}_{n\text{-times}}$
denotes a repeated application of a convolution mapping $G_k$ defined by
an averaging filter of size $k \times k$ with all entries set to $1 / k^2$.
We treat the numbers $n, k \in \mathbb{N}$, $k \geqslant 1$ as hyperparameters,
where $G^0_k$ is the identity mapping.
\begin{figure}[t]
\centering
\includegraphics[scale = 0.95]{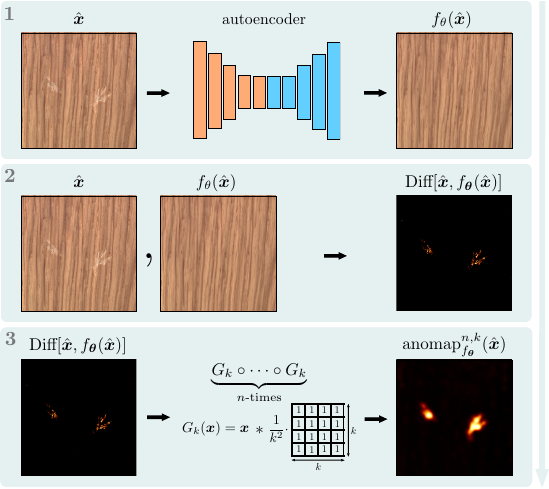}
\caption{
Illustration of our AD process.
Given input $\hat{\bfx}$, we first compute an output $f_{\bftheta}(\hat{\bfx})$
by replicating normal regions and replacing irregularities with locally consistent patterns.
Then we compute a pixel-wise squared difference $(\hat{\bfx} - f_{\bftheta}(\hat{\bfx}))^2$,
which is subsequently averaged over the color channels to produce the difference map $\text{Diff}[\hat{\bfx}, f_{\bftheta}(\hat{\bfx})] \in \mathbb{R}^{h \times w}$.
In the last step we apply a series of averaging convolutions $G_k$ to the difference map to produce our final anomaly heatmap $\text{anomap}_{f_{\bftheta}}^{n,k}(\hat{\bfx})$.
}
\label{fig_bauer3}
\end{figure}
By thresholding $\text{anomap}_{f_{\bftheta}}^{n,k}(\hat{\bfx})$ we get a binary segmentation mask for the anomalous regions.
We compute the anomaly score for the entire image $\hat{\bfx}$ from $\text{anomap}_{f_{\bftheta}}^{n,k}(\hat{\bfx})$
by summing the scores for the individual pixels\footnote{
Note that for $n = 0$, if we skip the averaging step over the color-channels in (\ref{eq_diffmap}), this reduces to $\|\hat{\bfx} - f_{\theta}(\hat{\bfx})\|_2^2$.
Alternatively to the summation, we could take the maximum over the pixel scores making the anomaly score potentially less sensitive to size variation of the anomalous regions.}.
The complete detection procedure is summarized in Figure \ref{fig_bauer3}.
In the next section, we investigate which image transformations approximately preserve the orthogonality of the corresponding transformations.

\subsection{Input Modifications}
\label{sec:gen}
\begin{figure*}[t!]
\centering
\includegraphics[scale = 0.8]{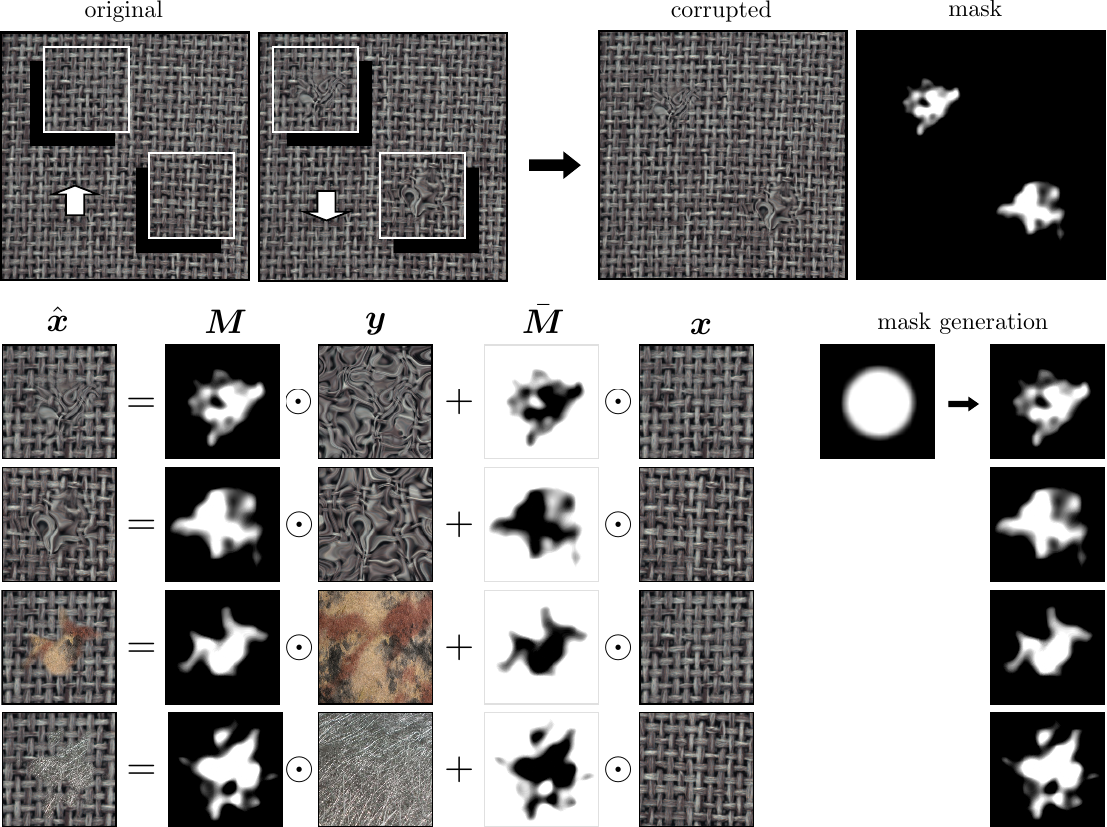}
\caption{
Illustration of data generation for training.
After randomly choosing the locations of the patches to be modified, we create a new content by glueing the extracted patches
with the corresponding replacements. Given a real-valued mask $\bfM \in [0,1]^{\tilde{h} \times \tilde{w} \times 3}$ marking corrupted regions within a patch,
an original image patch $\bfx$, and a corresponding replacement $\bfy$,
we create the next corrupted patch by merging the two patches together according to the formula $\hat{\bfx} := \bfM \odot \bfy + \bar{\bfM} \odot \bfx$.
All mask shapes $\bfM$ are created by applying gaussian distortion to the same (static) mask representing a filled disk at the center of the patch with a smoothly fading boundary towards the exterior of the disk.
}
\label{fig_bauer4}
\end{figure*}

Since we use a self-supervised approach we need to provide the training data as input-output $(\hat{\bfx}, \bfx)$ pairs.
Here, the ground truth outputs are given by the original images $\bfx$ corresponding to the normal examples.
The inputs $\hat{\bfx}$ are generated from these images by partially modifying some regions
according to the procedure illustrated in Figure \ref{fig_bauer4}.

For each normal example $\bfx$, we first randomly sample the number, the size and the location of patches to be modified.
In the next step, each randomly selected patch is modified according to the following procedure.
First, we create a real-valued mask $\bfM \in [0, 1]^{\tilde{h} \times \tilde{w} \times 3}$
based on the elastic deformation technique to mark the corrupted regions within the selected patch.
Precisely, we start with a static mask shaped as a disk at the center of the patch with a smoothly fading boundary towards the exterior of the disk.
We then apply gaussian distortion (with random parameters) to this mask resulting in varying shapes illustrated on the right side in Figure \ref{fig_bauer4}.
These masks are used to smoothly merge the original patches with a new content given by replacement patches.
Here, we consider two types of replacements. On the one hand we can use any natural image (sufficiently different from the original patch) as a potential replacement.
In our experiments, we used the publicly available Describable Textures Dataset (DTD) \cite{cimpoi14describing} consisting of images of varying backgrounds.
On the other hand, we use the extracted patches as replacements
after passing them through a gaussian distortion process similarly to how we create the masking shapes.
An important aspect here is that the corresponding corruptions approximately preserve the original color distribution.

Given an input $\bfx$ and a replacement $\bfy$, we create a corrupted image $\hat{\bfx}$
by smoothly gluing the two images together according to the formula $\hat{\bfx} := \bfM \odot \bfy + \bar{\bfM} \odot \bfx$.
In the last step, the patches extracted at the beginning are replaced by their modified versions in the original image.
The individual shape masks are embedded in a two-dimensional zero-array at the corresponding locations to create a global mask $\bfM \in [0, 1]^{h \times w \times 3}$.
During testing phase there are no input modifications and images are passed unchanged to the network.
Anomalous regions are then detected by thresholding the anomaly heatmap based on the difference
between the inputs and the outputs of the model according to the formula in equation (\ref{eq_anomap}).


\subsection{Mode Architecture}
\label{sec_model}
We follow an established architecture of deep auto-encoders,
where the encoder is build from layers gradually decreasing spatial dimension
and increasing the number of feature maps,
while the decoder reverses this process by increasing the spatial dimension and decreasing the number of channels.
Additionally to the standard convolutions, we exploit the dilated convolutions \cite{SchusterWUS19} for approximating higher order dependencies
between the individual pixels and their local neighbourhood.
\begin{table}[t] 
  \scriptsize
  \caption{Network architecture Model I.}
  \label{table_model}
  \centering
  \begin{tabular}{lcclc}
    \toprule
    Layer Type     &  Number of Filters  &  Filter Size  &  Output Size  \\
    \midrule
    Conv &  $64$   &  $3 \times 3$ & $512 \times 512 \times 64$\\
    Conv &  $64$   &  $3 \times 3$ & $512 \times 512 \times 64$\\
    MaxPool &              &  $2 \times 2$ / $2$ & $256 \times 256 \times 64$\\
    		       &		&		  & \\
    Conv &  $128$   &  $3 \times 3$ & $256 \times 256 \times 128$\\
    Conv &  $128$   &  $3 \times 3$ & $256 \times 256 \times 128$\\
    MaxPool &             &  $2 \times 2$ / $2$ & $128 \times 128 \times 128$\\
    		       &	     &                                & \\		
    Conv &  $256$ &  $3 \times 3$ & $128 \times 128 \times 256$\\
    Conv &  $256$ &  $3 \times 3$ & $128 \times 128 \times 256$\\
    MaxPool &             &  $2 \times 2$ / $2$ & $64 \times 64 \times 256$\\
    		       &             &                                & \\
    SDC$_{1, 2, 4, 8, 16, 32}$ &  $64 \times 6$  &  $3 \times 3$ & $64 \times 64 \times 384$\\
    SDC$_{1, 2, 4, 8, 16, 32}$ &  $64 \times 6$  &  $3 \times 3$ & $64 \times 64 \times 384$\\
    SDC$_{1, 2, 4, 8, 16, 32}$ &  $64 \times 6$  &  $3 \times 3$ & $64 \times 64 \times 384$\\
    SDC$_{1, 2, 4, 8, 16, 32}$ &  $64 \times 6$  &  $3 \times 3$ & $64 \times 64 \times 384$\\
    		       &             &                                & \\
    TranConv & $256$ &  $3 \times 3$ / $2$  & $128 \times 128 \times 256$\\
    Conv & $256$ &  $3 \times 3$ & $128 \times 128 \times 256$\\
    Conv & $256$ &  $3 \times 3$ & $128 \times 128 \times 256$\\
    		       &             &                                & \\
    TranConv &  $128$ &  $3 \times 3$ / $2$  & $256 \times 256 \times 128$\\
    Conv &  $128$ &  $3 \times 3$ & $256 \times 256 \times 128$\\
    Conv &  $128$ &    $3 \times 3$ & $256 \times 256 \times 128$\\
    		       &             &                                & \\
    TranConv &  $64$ &  $3 \times 3$ / $2$  & $512 \times 512 \times 64$\\
    Conv &  $64$ &  $3 \times 3$ & $512 \times 512 \times 64$\\
    Conv &  $64$ &  $3 \times 3$ & $512 \times 512 \times 64$\\
    
    Conv &  $3$ &  $1 \times 1$ & $512 \times 512 \times 3$\\
    \bottomrule
  \end{tabular}
\end{table}
An overview of the whole architecture is given in Table \ref{table_model}.
Here, after each convolution \textit{Conv} (except in the last layer) we use batch normalization and rectified linear units (ReLU) as the activation function.
Max-pooling \textit{MaxPool} is applied to reduce the spatial dimension of the intermediate feature maps.
\textit{TranConv} denotes the convolution transpose operation and also uses batch normalization with ReLUs.
In the last layer we use a sigmoid activation function without batch normalization.
SDC$_{1, 2, 4, 8, 16, 32}$ refers to a stacked dilated convolution block where multiple dilated convolutions are stacked together.
The corresponding subscript $\{1, 2, 4, 8, 16, 32\}$ denotes the dilation rates of the six individual convolutions in each stack.
After each stack we add an additional convolutional layer with the kernel size $3 \times 3$ and the same number of feature maps as in the stack
followed by a batch normalization and ReLU activation.
We refer to this baseline architecture as Model I.
\begin{figure}[b!]
\centering
\includegraphics[scale = 0.82]{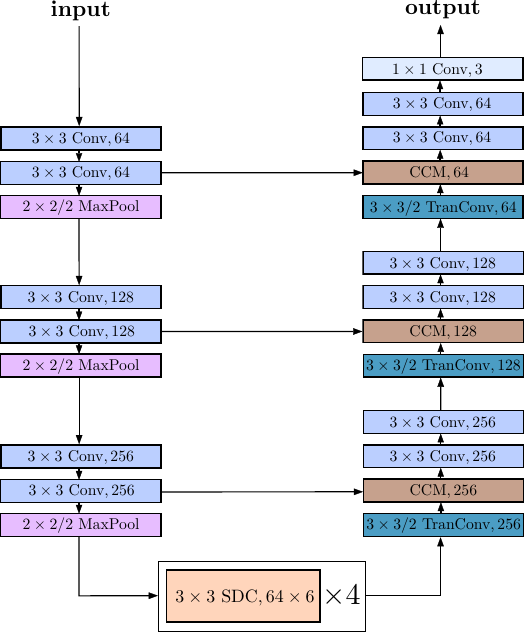}
\caption{
Illustration of our network architecture Model II including the Convex Combination Module (CCM)
marked with brown color and the skip-connections represented by the horizontal arrows.
Without these additional elements we get the baseline Model I.
}
\label{fig_bauer6}
\end{figure}

\begin{figure}[t]
\centering
\includegraphics[scale = 0.6]{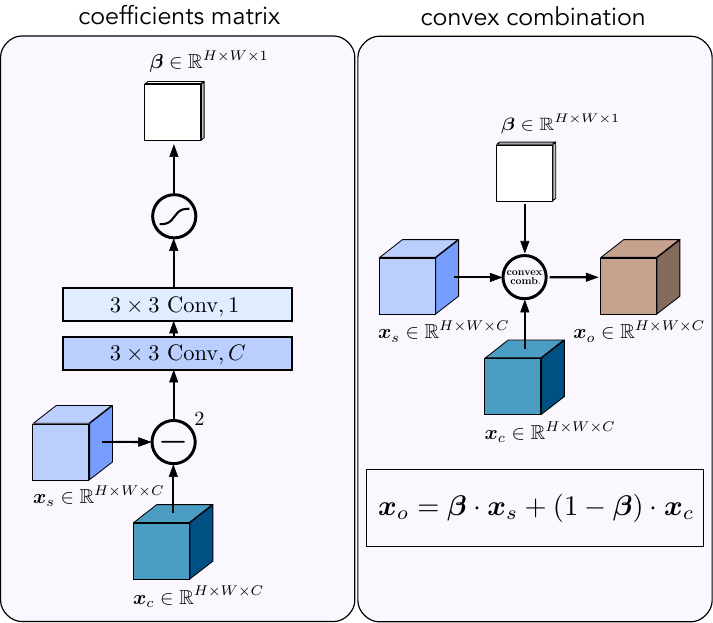}
\caption{
Illustration of the CCM module.
The module receives two inputs: $\bfx_s$ along the skip connection and $\bfx_c$ from the current layer below.
In the first step (image on the left) we compute the squared difference of the two and stack it together with
the original values $[\bfx_s, \bfx_c, (\bfx_s - \bfx_c)^2]$. This combined feature map is processed
by two convolutional layers. The first layer uses batch normalization with ReLU activation.
The second layer uses batch normalization and sigmoid activation function
to produce a coefficient matrix $\bfbeta$.
In the second step (image on the right), we compute the output of the module $\bfx_o$
as a convex combination of the inputs by broadcasting $\bfbeta$ along the last dimension.
}
\label{fig_bauer7}
\end{figure}

In our experiments,
we observed that Model I struggles to reproduce finer visual patterns sometimes resulting in false positive detection.
In order to improve the reconstruction ability, we propose the following adjustment illustrated in Figure \ref{fig_bauer6}.
Similarly to the idea often used in image segmentation, we introduce skip connections into the network.
Providing direct access to feature maps from the earlier stages seems (sometimes) to encourage the network to neglect other parts of the computation responsible for the reconstruction of corrupted regions. 
To prevent this, we combine the information from different layers in the network by leveraging some sort of attention mechanism.
Precisely, we first compute a matrix of coefficients between zero and one,
which we then use to compute a convex combination of the individual feature maps.
The corresponding procedure is illustrated in Figure \ref{fig_bauer7}.
The intuition here is that the network now can learn to reuse, refine or replace
regions from the input based on its belief about the abnormality of a region.
We bundle the corresponding computations in a single Convex Combination Module (CCM)
marked brown in Figure \ref{fig_bauer6}.
We refer to the corresponding architecture as Model II.
In our experiments on the MVTec AD dataset,
we observed a significant performance boost on some object categories when using Model II over I.
See Figure \ref{fig_bauer14} in the supplements for comparison of the reconstruction quality of both models.
\section{Theoretical Analysis}
\label{sec:section4}
Here we provide a theoretical analysis supporting the intuition behind the AD method proposed in the previous section.
In particular, we show that the resulting model approximates the orthogonal projection
of partially corrupted images onto the submanifold of normal samples.
We begin by establishing a close connection to the previous regularized AEs by identifying the orthogonal
projection as an optimal solution for the optimization problem of the RCAE.
We then investigate the conservation properties of the orthogonal projections
with respect to the segmentation accuracy in the context of AD.
Finally, we specify the form of input modifications ensuring the orthogonality of the corresponding transformations.

\begin{figure}[t]
\centering
\includegraphics[scale = 0.5]{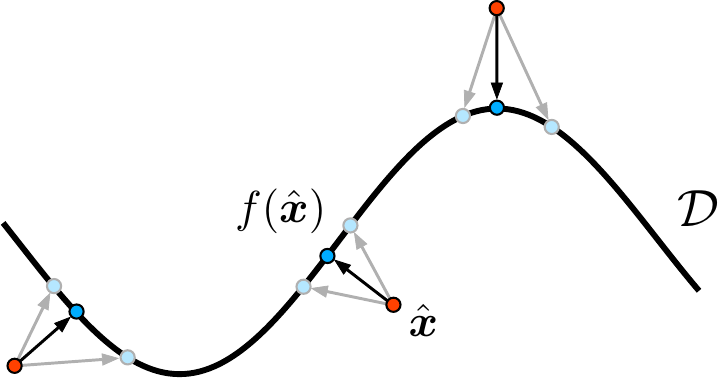}
\caption{
Illustration of the general concept of the orthogonal projection $f$ onto a data manifold $\mathcal{D}$.
Here, anomalous samples $\hat{\bfx} \in \mathbb{R}^n$ (red dots) are projected to points $\bfx := f(\hat{\bfx}) \in \mathcal{D}$
(blue dots) in a way minimizing the distance $d(\hat{\bfx}, \bfx) = \inf_{\bfy \in \mathcal{D}} d(\hat{\bfx}, \bfy)$.
}
\label{fig_bauer9}
\end{figure}

For the purpose of the following analysis,
we identify each projecting autoencoder (PAE) with an idempotent mapping
$f \colon \mathcal{U} \rightarrow \mathbb{R}^n$
from an input space $\mathcal{U} \subseteq \mathbb{R}^n$, $n \in \mathbb{N}$
to a differentiable manifold $\mathcal{D} := f(\mathcal{U}) \subseteq \mathbb{R}^n$, $\textit{dim}(\mathcal{D}) < n$.
In the context of AD, we refer to the manifold $\mathcal{D}$
as the (non-linear) subspace of normal examples,
while each $\hat{\bfx} \in \mathbb{R}^n \setminus \mathcal{D}$ is considered to be anomalous.
In particular, we exploit a generalization of orthogonal projection to non-linear mappings defined below and illustrated in Figure \ref{fig_bauer9}.
\begin{definition}
\begin{itshape}
For $n \in \mathbb{N}$ in a metric space $(\mathbb{R}^n , d(\cdot, \cdot))$,
 we call a (non-linear) mapping $f \colon \mathcal{U} \rightarrow \mathbb{R}^n$, $\mathcal{U} \subseteq \mathbb{R}^n$ the \textbf{orthogonal projection} 
onto $\mathcal{D} \subseteq \mathbb{R}^n$ if it satisfies the equality
\begin{equation}
\label{E_08040853}
d(\hat{\bfx}, f(\hat{\bfx})) = \inf_{\bfy \in \mathcal{D}} d(\hat{\bfx}, \bfy)
\end{equation}
for all $\hat{\bfx} \in \mathcal{U}$.
\end{itshape}
\end{definition}
Note that the expression in (\ref{E_08040853}) can be written as a set of inequalities.
Namely, for $d(\hat{\bfx}, f(\hat{\bfx})) := \|\hat{\bfx} - f(\hat{\bfx})\|_2$, we get the following alternative description
\begin{equation}
\label{E_09041757}
\forall \bfy \in \mathcal{D} \hspace*{1pt} \colon \hspace*{3pt} \| \hat{\bfx} - f(\hat{\bfx}) \|_2 \hspace*{2pt}\leqslant \| \hat{\bfx} - \bfy \|_2.
\end{equation}
We will use the form (\ref{E_09041757}) later in Section \ref{subsec:data_aug}.

In order to train a PAE, we minimize the following objective
\footnote{
Compare to the objective in (\ref{E_12041904}).
There we explicitly use a mask $\bfM$ only to balance the training with respect to the reconstruction accuracy of normal and anomalous regions.
If we fix the value of $\|\bfM\|_1$ and set $\lambda = \|\bfM\|_1 / (\|\bfM\|_1 + \|\bar{\bfM}\|_1)$,
the objective (\ref{E_12041904}) reduces to minimizing 
the loss
$\mathcal{L}(\hat{\bfx}, \bfx) = \|f_{\bftheta}(\hat{\bfx}) - \bfx\|_2^2$.
}
\begin{equation}
\label{E_14031525}
\mathcal{L}_{PAE} = \mathbf{E}\left[\|f_{\bftheta}(T(\bfx)) - \bfx\|_p^p\right],
\end{equation}
where $p \in \mathbb{N}_+$
and $T \colon \mathcal{D} \rightarrow \mathbb{R}^n$ denotes some random data transformation.
That is, the expectation in (\ref{E_14031525}) is taken with respect to $\bfx \in \mathcal{D}$ and $T \in \mathcal{T}$.
Note that the shape of $\mathcal{T}$ mainly determines the properties of the projection map to be learned,
which, in turn, can be seen as the reverse mapping of the corresponding input corruptions $T$.
Depending on our goal, $\mathcal{T}$ can be very specific
ranging from additive noise (e.g. gaussian noise)
to partial occlusions and elastic deformations.
In Section \ref{subsec:data_aug}, we identify a number of input transformations,
which approximately preserve the orthogonality of the corresponding projections.

\subsection{Connections to Regularized Autoencoders}
Typically, an autoencoder is composed from two building blocks:
the encoder $e(\cdot)$, which maps the input $\bfx$ to some internal representation,
and the decoder $d(\cdot)$, which maps $e(\bfx)$ back to the input space.
The composition $f(\bfx) = d(e(\bfx))$ is often referred to as the reconstruction mapping.
Most of the regularized autoencoders aim to capture the structure of the training distribution
based on an interplay between the reconstruction error and the regularization term.
They are trained by minimizing the reconstruction loss on the training data
either by directly adding a regularization term to the objective
or by introducing the regularization through some kind of data augmentation.

Specifically, the contractive autoencoder (CAE) \cite{RifaiVMGB11} is trained to minimize the following regularized reconstruction loss
\begin{equation}
\mathcal{L}_{CAE} = \mathbf{E}\left[\|f(\bfx) - \bfx\|_2^2 + \lambda \|D_{\bfx}e\|^2_F\right],
\end{equation}
where $\lambda \in \mathbb{R}_+$ is a weighting hyperparameter and $\|D_{\bfx}e\|_F$ is the Frobenius norm of the Jacobian of the encoder $e(\cdot)$.
The denoising autoencoder (DAE) \cite{VincentLBM08}, on the other hand, is trained to minimize the following denoising criterion
\begin{equation}
\mathcal{L}_{DAE} = \mathbf{E}\left[\|f(\bfx + \epsilon) - \bfx\|_2^2 \right],
\end{equation}
where $\epsilon$ represents some additive noise and the expectation is over the training distribution and the corruption noise
\footnote{Here, the term noise includes additive (e.g., isotropic gaussian noise) and non-additive transformation (e.g., masking pepper-noise).
However, the unifying feature of the considered corruptions is that the transformations of the individual entries in the input are statistically independent.
In contrast, we consider in (\ref{E_14031525}) more complex modifications, which result in strong spatial correlation of the corruption variables.
}.
Previous work \cite{AlainB14} has demonstrated that there is close connection between the CAE and DAE
when the standard deviation of the corruption noise approaches zero $\sigma \rightarrow 0$.
More precisely, they showed that under some technical assumptions
the objective of the DAE can be written as
\begin{equation}
\mathcal{L}_{DAE} = \mathbf{E}\left[\|f(\bfx) - \bfx\|^2_2 + \sigma^2 \|D_{\bfx}f\|^2_F\right] \text{ as } \sigma \rightarrow 0.
\end{equation}
That is, for small variance $\sigma^2$ of the corruption noise
the DAE  becomes similar to a CAE with penalty coefficient $\lambda = \sigma^2$,
but where the contraction is imposed explicitly on the whole reconstruction mapping $f$.
This connection motivated the authors to define the RCAE, a variation of the CAE
by minimizing the following objective
\begin{equation}
\label{E_14030956}
\mathcal{L}_{RCAE} = \mathbf{E}\left[\|f(\bfx) - \bfx\|^2_2 + \lambda \|D_{\bfx}f\|^2_F\right],
\end{equation}
where the regularization term affects the whole reconstruction mapping.
Finally, a common variant of the SAE applies an $l^1$-penalty on the hidden unit activations
effectively making them saturate towards zero -- similarly to the effect in the CAE.

\begin{figure}[b]
\centering
\includegraphics[scale = 0.55]{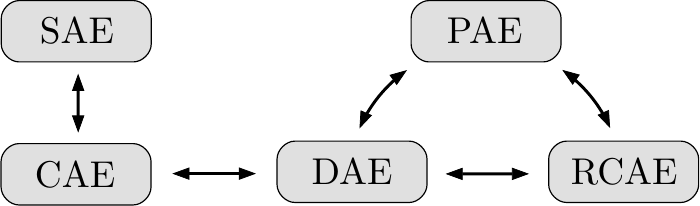}
\caption{
Illustration of the connections between the different types of regularized autoencoders.
For small variance of the corruption noise, the DAE
becomes similar to the CAE.
This, in turn, gave rise to the RCAE,
where the contraction is imposed explicitly on the whole reconstruction mapping.
A special instance of PAE given by the orthogonal projection yields an optimal solution
for the optimization problem of the RCAE. On the other hand, the training objective
for PAE can be seen as an extension of DAE to more complex input modifications beyond additive noise.
Finally, a common variant of the SAE applies an $l^1$-penalty on the hidden units
 resulting in saturation towards zero similarly to the CAE.
}
\label{fig_bauer10}
\end{figure}

We now show that a PAE realized by the orthogonal projection onto the submanifold
of normal samples
is an optimal solution minimizing the objective of the RCAE in (\ref{E_14030956}).
We provide a formal proof in the supplements.
\begin{proposition}
\label{prop_14031216}
\begin{itshape}
Let $f^* \colon \mathcal{U} \subseteq \mathbb{R}^n \rightarrow \mathbb{R}^n, n \in \mathbb{N}$
be an orthogonal projection onto $\mathcal{D} := f^*(\mathcal{U})$ with respect to an $l^p$-norm, $p \in \mathbb{N}_+$.
Then $f^*$ is an optimal solution to the following optimization problem
\end{itshape}
\begin{equation}
\label{E_08040946}
\underset{f \in C^1}{\text{minimize}} \hspace*{5pt} \mathbf{E}_{\bfx \sim \mathcal{D}}\left[\|f(\bfx) - \bfx\|^2_2 + \lambda \left\|D_{\bfx}f\right\|^2_q \right],
\end{equation}
\begin{itshape}
where $\lambda \in \mathbb{R}_+$ and $q \in \{F, 2\}$ is a placeholder denoting either the Frobenius or the spectral norm.
\end{itshape}
\end{proposition}
Note that the objective in the above proposition is slightly more general than in (\ref{E_14030956}) allowing the use of the spectral norm.
Although being closely related to each other, the Frobenius and the spectral norm might have different effects on the optimization.
Namely, the spectral norm measures the maximal scale by which a unit vector is stretched by a corresponding linear transformation
and is determined by the maximal singular value,
while the Frobenius norm measures the overall distortion of the unit circle taking into account all the singular values.
In a more general sense, Proposition \ref{prop_14031216} reveals the close connection between the PAE
and other types of regularized autoencoders (see Figure \ref{fig_bauer10} for an overview).
The orthogonal projection, in particular, appears to be an optimal choice\footnote{
In Section \ref{subsubsec:add_noise}, we will show that in the limit when the dimensionality of the inputs goes to infinity,
the DAE converges stochastically to the orthogonal projection restricted to the specific noise corruptions.
}
with respect to the shared goals of the autoencoding models discussed in this section.

\subsection{Conservation Effect of Orthogonal Projections}
\label{sec:section4.2}
Here, we focus on the subtask of AD regarding the segmentation of anomalous regions in the inputs.
In the following, we identify each image tensor with a column vector $\bfx \in \mathbb{R}^n$,
where $S \subset \{1, ..., n\} $ and $\bar{S} := \{1, ..., n\} \setminus S$
denote a set of pixel indices and its complement, respectively.
We write $\bfx_S$ to denote a restriction of a vector $\bfx$ to the indices in $S$.

Consider an $\hat{\bfx} \in \mathbb{R}^{n}$, which has been generated by a (partial) modification $\hat{\bfx} := T(\bfx)$, $T \in \mathcal{T}$ from an $\bfx \in \mathcal{D}$.
We define the \textbf{modified region} as
the set of disagreeing indices according to
\begin{equation}
S(\hat{\bfx}, \bfx) := \left\{i \in \{1, ..., n\} \colon \hat{x}_i \neq x_i \right\}.
\end{equation}
There is some ambiguity about what might be considered an anomalous region,
which motivates the following definition.
\begin{definition}
\begin{itshape}
Let $\mathcal{D} \in \mathbb{R}^n$, $n \in \mathbb{N}$ be a data manifold. Given $\hat{\bfx} \in \mathbb{R}^n \setminus \mathcal{D}$,
we define the set of anomalous regions in $\hat{\bfx}$ as the areas of smallest disagreement according to
\begin{equation}
\label{E_S_A}
\mathcal{A}(\hat{\bfx}) := \left\{S(\hat{\bfx}, \bfx) \hspace*{3pt}\colon\hspace*{3pt} \bfx \in \argmin_{\bfy \in \mathcal{D}} |S(\hat{\bfx}, \bfy)| \right\}.
\end{equation}
 We refer to each $S \in \mathcal{A}(\hat{\bfx})$ as \textbf{anomalous region}, to the pair $(S, \hat{\bfx}_{S})$ as \textbf{anomalous pattern}
and to $\hat{\bfx}$ as \textbf{anomalous sample}.
\end{itshape}
\end{definition}
We now provide some intuition motivating our definition of anomalous patterns.
Consider an example of binary sequences $\bfx \in \{0, 1\}^8$ restricted by the condition that the pattern "11" is forbidden.
For example, the sequence $(0,1,0,\redOn1\blackOn,\redOn1\blackOn,\redOn1\blackOn,\redOn1\blackOn,0)$ is invalid because it contains "11".
If we define the anomalous region as the smallest subset of indices, which need to be corrected
in order for the point to be projected to the data manifold, we encounter the following ambiguity problem.
Namely, there are three different ways to correct the above example using minimal number of changes.
\begin{figure}[t]
\centering
\includegraphics[scale = 0.85]{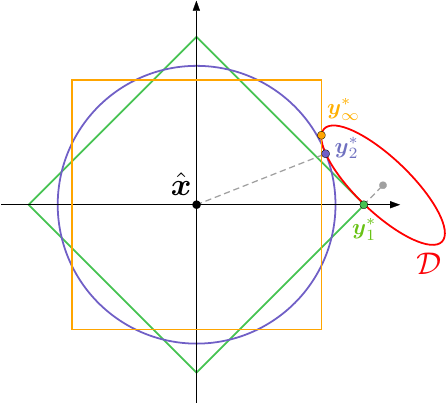}
\caption{
Illustration of the conservation effect of the orthogonal projections with respect to different $l^p$-norms.
Here, the anomalous sample $\hat{\bfx}$ is orthogonally projected onto the manifold $\mathcal{D}$
according to $\|\hat{\bfx} - \bfy^*_p\|_p = \inf_{\bfy \in \mathcal{D}} \|\hat{\bfx} - \bfy\|_p$
for $p \in \{1, 2, \infty\}$. We can see that projections $\bfy^*_p$ for lower $p$-values better preserve the content in $\hat{\bfx}$
resulting in smaller modified regions $S(\hat{\bfx}, \bfy^*_p)$.
}
\label{fig_bauer11}
\end{figure}
These are
$(0,1,0, 1,\blueOn0\blackOn,1,\blueOn0\blackOn, 0)$,
$(0,1,0,\blueOn0\blackOn,1,\blueOn0\blackOn,1, 0)$,
and $(0,1,0, 1,\blueOn0\blackOn,\blueOn0\blackOn,1, 0)$,
which we denote as $\bfy_1$, $\bfy_2$ and $\bfy_3$, respectively.
All three sequences correspond to orthogonal projections to the feasible set,
where $\| \hat{\bfx} - \bfy_1 \|_2 = \| \hat{\bfx} - \bfy_2 \|_2 = \| \hat{\bfx} - \bfy_3 \|_2 = \sqrt{2}$.
However,
\begin{equation}
\bigcap_{i=1}^3 S(\hat{\bfx}, \bfy_i) \neq \bigcup_{i=1}^3 S(\hat{\bfx}, \bfy_i).
\end{equation}
That is, there are multiple anomalous patterns giving rise to the same corrupted point $\hat{\bfx}$.
In particular, the anomalous regions in $\hat{\bfx}$ are not uniquely determined and depend on the structure of $\mathcal{D}$.
Based on this terminology, we highlight below a special projection map,
which maximally preserves the content of its arguments.
\begin{definition}
\label{def_cons}
\begin{itshape}
For $n \in \mathbb{N}$, we call a mapping $f \colon \mathcal{U} \rightarrow \mathbb{R}^n$, $\mathcal{U} \subseteq \mathbb{R}^n$
the \textbf{conservative projection} 
onto $\mathcal{D} \subseteq \mathbb{R}^n$ if for each pair $(\hat{\bfx}, \bfx)$ with $f(\hat{\bfx}) = \bfx$ and $S := S(\hat{\bfx}, \bfx)$ it satisfies the following properties
\begin{equation*}
\begin{aligned}
(a)& \hspace*{10pt} S \in \mathcal{A}(\hat{\bfx})\\
(b)& \hspace*{10pt} \|\hat{\bfx}_S - f_S(\hat{\bfx})\|_2 \hspace*{3pt} \leqslant \hspace*{1pt}{\inf}_{\bfx \in \mathcal{D} \hspace*{1pt} \colon \hspace*{1pt} \hat{\bfx}_{\bar{S}} = \bfx_{\bar{S}}} \hspace*{1pt}\|\hat{\bfx}_S - \bfx_S\|_2.
\end{aligned}
\end{equation*}
\end{itshape}
\end{definition}
The next proposition relates the conservation properties of orthogonal projections for different $l^p$-norms.
See Figure \ref{fig_bauer11} for an illustration.
\begin{proposition}
\label{prop_14031813}
\begin{itshape}
Let $\mathcal{D} \subseteq \mathbb{R}^n$, $n \in \mathbb{N}$ be a data manifold and $\hat{\bfx} \in \mathbb{R}^n \setminus \mathcal{D}$ a
corrupted example.
Furthermore, let $S \in \mathcal{A}(\hat{\bfx})$ and $S_p := S(\hat{\bfx}, \bfy^*_p)$, where 
$\bfy^*_p \in \inf_{\bfy \in \mathcal{D}} \|\hat{\bfx} - \bfy\|_p$ corresponds to an orthogonal projection of $\hat{\bfx}$ onto $\mathcal{D}$ with respect to the $l^p$-norm.
For all $p, q \in \mathbb{N}_+$, $p < q$ the following statements are true:
\begin{equation*}
\begin{aligned}
(a)& \hspace*{10pt} S \subseteq S_{p} \subseteq S_{q}\\
(b)& \hspace*{10pt} \exists \mathcal{D} \subseteq \mathbb{R}^n \colon \hspace*{3pt} S_{1} \neq S_{2}.
\end{aligned}
\end{equation*}
\end{itshape}
\end{proposition}
The above proposition shows that orthogonal projection with respect to the $l^p$-norm is more conservative for lower $p \geqslant 2$ values
regarding the preservation of normal regions\footnote{
However, $l^2$-norm has practical advantages over $l^1$-norm
regarding the optimization process and is, therefore, often a better choice.}.
Furthermore, the orthogonal projection with respect to the $l^2$-norm (unlike the conservative projection) is not maximally preserving in general.
We will show later, however, that in the limit (when the dimensionality of the input vectors goes to infinity)
the conservative and the $l^2$-orthogonal projection disagree only on a zero-measure probability set.

While Proposition \ref{prop_14031813} describes the conservation properties of orthogonal projections relative to each other,
the following proposition specifies (asymptotically) how accurate the corresponding reconstruction error describes the anomalous region.
Here, we introduce the notion of a transition set, which glues the disconnected parts of an input together illustrated in Figure \ref{fig_bauer12}.
We provide more details in the supplements.
\begin{proposition}
\label{P_09041241}
\begin{itshape}
Let $f \colon [0, 1]^n \rightarrow \mathcal{D} \subseteq [0, 1]^n$, $n \in \mathbb{N}$ be the orthogonal projection
with respect to an $l^p$-norm, $p \in \mathbb{N}_+$ and  $\bfx \in \mathcal{D}$ with a finite set of states $x_i \in I \subseteq [0, 1]$, $|I| < \infty$.
Consider an $\hat{\bfx} \in [0, 1]^n$, which has been generated from $\bfx$ by partial modification with
$\hat{\bfx}_{\bar{S}} = \bfx_{\bar{S}}$ for some $S \subset \{1, ..., n\}$, $\bar{S} := \{1, ..., n\} \setminus S$.
Furthermore, let $B \subseteq \{1, ..., n\}$ denote a transition set from $S$ to $\bar{S} \setminus B$.
For all $p \in \mathbb{N}$, $p \geqslant 1$ the following holds true:
\begin{equation}
\label{E_14031900}
\|f_{\bar{S}}(\hat{\bfx}) - \bfx_{\bar{S}}\|_p \leqslant |B|^{\frac{1}{p}},
\end{equation}
where $|B|$ grows asymptotically according to $O(\sqrt{|S|})$.
\end{itshape}
\end{proposition}
When projecting corrupted points onto the data manifold,
we would like the transition set $B$ around the anomalous region $S$ to be as small as possible.
The inequality in (\ref{E_14031900}) implicitly upper-bounds the number of entries in the normal region $\bar{S}$,
which are not preserved by the projection.
It is mostly practical for lower values of $p$ and least informative\footnote{
Namely, $\underset{n \hspace*{2pt}\rightarrow \hspace*{2pt} \infty}{\text{lim}} |B|^{\frac{1}{p}} = 1$ provides a trivial upper bound,
since $\|\bfx\|_{\infty} \leqslant 1$ for $\bfx \in [0, 1]^n$.
}
in the case $p = \infty$.
For $p = 1$, for example, the interpretation is the simplest in the case of short sequences of binary values.
The number of disagreeing components is then upper-bounded by the number $|B|$
corresponding to the size of the transition set, which, in turn, is determined by the form of the underlying data distribution.
\begin{figure}[t]
\centering
\includegraphics[scale = 0.52]{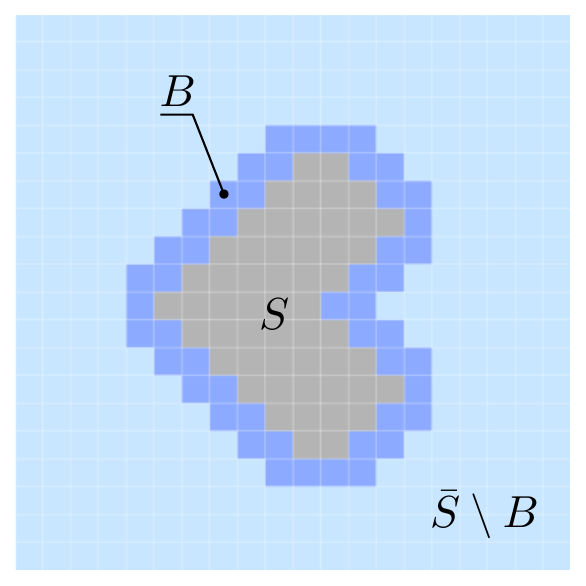}
\caption{
Illustration of the concept of a transition set. Consider a 2d-image tensor identified with a column vector $\bfx \in \mathbb{R}^n$, $n = 20^2$,
which is partitioned according to $S \subseteq \{1, ..., n\}$ (gray area) and $\bar{S} := \{1, ..., n\} \setminus S$ (union of light-blue and dark blue areas).
The transition set $B$ (dark blue area) glues the two disconnected sets $S$ and $\bar{S}\setminus B$ together such that $\bfx$ is feasible.
}
\label{fig_bauer12}
\end{figure}

To summarize, we showed that orthogonal projection with respect to the $l^p$-norm
preserves the normal regions more accurately for smaller $p$-values.
Furthermore, we identified the conservative projection as the one, which is maximally preserving (unlike the orthogonal projection).
As previously mentioned, we will show in Section \ref{subsec:data_aug} that in the limit (when the dimensionality of the input vectors goes to infinity)
the conservative and the $l^2$-orthogonal projection are the same up until a zero-measure probability set.

\subsection{Data Augmentation}
\label{subsec:data_aug}
In the following, we specify a range of input modifications, which (approximately)
preserve the orthogonality of the corresponding projections.
In the context of image processing, the plethora of existing data augmentation techniques can be roughly divided into five groups
corresponding to the affine transformations, color jittering, mixing strategies, elastic deformations, and additive noise.
Here, we characterize each image transformation
either as $(a)$ a partial modification (of any type) or as $(b)$ a modification affecting the entire image
represented by the additive noise methods.
Image mixing strategies like MixUp, for example, can simply be seen as a transformation with additive noise,
while CutMix is an example of partial modification.
On the other hand, linear and affine data transformations like shift, rotation or color-channel permutation,
in general, do not preserve orthogonality.

For the purpose of the subsequent analysis,
we identify the image tensors in our input space with multivariate random variable $\bfx \in \mathbb{R}^n$
corresponding to Markov Random Fields (MRFs)\cite{Koller:2009:PGM:1795555, WainwrightJ08, Lafferty01conditionalrandom, BauerSM17, Bauer2019, math11122628}.
In particular, we assume that the variables representing the individual pixels
are organized in a two-dimensional grid.
Based on this view,
we consider sequences $\mathcal{D}^{(n)} \subseteq [0, 1]^n$
of spaces  of increasing dimensionality $n \in \mathbb{N}$
representing images of gradually increasing size by adding new nodes along the rows and columns of the grid.
Furthermore, we use the notation $d_G(x_i, x_j)$ to denote the distance
between the variables $x_i$ and $x_j$ in a corresponding MRF graph $G$
defined by the number of edges on a shortest path connecting the two nodes.

\subsubsection{Partial Modification}
The following theorem describes the technical conditions, under which 
a corresponding partial modification (of arbitrary type) corresponds to the orthogonal projection.
\begin{theorem}
\label{theorem_1}
\begin{itshape}
Consider a pair $\bfx, \bfy \in \mathcal{D}^{(n)}$ of independent MRFs
over identically distributed variables $x_i, y_j$ with finite fourth moment $\mathbf{E}[x_i^4] < \infty$,
variance $\sigma^2 := \mathbf{Var}[x_i]$
and vanishing covariance
\begin{equation}
\mathbf{Cov}[x_i^k, x_j^l] \hspace*{2pt} \longrightarrow \hspace*{2pt} 0 \hspace*{5pt} \text{ for } \hspace*{5pt} d_G(x_i, x_j) \hspace*{2pt} \longrightarrow \hspace*{2pt} \infty
\end{equation}
for all $k, l \in \{1, 2\}$.
Furthermore, let $\hat{\bfx} \in [0, 1]^n$ be a copy of $\bfx$, which has been partially modified, where
$\hat{\bfx}_{\bar{S}} = \bfx_{\bar{S}}$ for some $S \subset \{1, ..., n\}$, $\bar{S} := \{1, ..., n\} \setminus S$.
The following is true
\begin{equation}
\label{E_convergence}
\underset{n \hspace*{2pt}\rightarrow \hspace*{2pt} \infty}{\text{lim}} \mathbb{P}_{\bfx, \bfy}\left(\| \hat{\bfx} - \bfx \|_2 \hspace*{2pt}\leqslant \| \hat{\bfx} - \bfy \|_2 \right) = 1,
\end{equation}
provided the inequality
\begin{equation}
\label{E_size}
|S| \hspace*{2pt}\leqslant 2 \cdot \sigma^2 \cdot |\bar{S}|
\end{equation}
holds for all $n \in \mathbb{N}$.
\end{itshape}
\end{theorem}
Given two samples $\bfx, \bfy \in \mathcal{D}$ and a corrupted copy $\hat{\bfx} \in [0, 1]^n$,
Theorem \ref{theorem_1} describes
how the expression $\| \hat{\bfx} - \bfx \|_2 \leqslant \| \hat{\bfx} - \bfy \|_2$
approaches a true statement when the dimensionality of the embedding space $n \in \mathbb{N}$ goes to infinity.
In particular, $\bfx$ can be identified with the image of $\hat{\bfx}$ under the conservative projection $f(\hat{\bfx}) = \bfx$.
That is,
\begin{equation}
\label{E_15041140}
\underset{n \hspace*{2pt}\rightarrow \hspace*{2pt} \infty}{\text{lim}} \mathbb{P}_{\bfx, \bfy}\left(\| \hat{\bfx} - f(\hat{\bfx}) \|_2 \hspace*{2pt}\leqslant \| \hat{\bfx} - \bfy \|_2 \right) = 1.
\end{equation}
Therefore, the conservative projection
converges stochastically to the orthogonal projection\footnote{
Compare to the definition of orthogonal projection in (\ref{E_09041757}).
},
while the inequality in (\ref{E_size}) controls the maximal size of corrupted regions $|S|$
by taking into account the distribution of normal examples.
On the other hand, the equation in (\ref{E_15041140}) suggests that
(in the limit) the orthogonal and the conservative projection
disagree at most on a subset of the data manifold $\mathcal{D}$, which has zero probability measure under the training distribution.

\subsubsection{Additive Noise}
\label{subsubsec:add_noise}
The following theorem describes the technical conditions, under which 
the denoising process (for arbitrary noise distribution) corresponds to the orthogonal projection.
\begin{theorem}
\label{theorem_2}
\begin{itshape}
Consider a pair $\bfx, \bfy \in \mathcal{D}^{(n)}$ of independent MRFs
over identically distributed variables $x_i, y_j$ with finite fourth moment $\mathbf{E}[x_i^4] < \infty$,
variance $\sigma^2 := \mathbf{Var}[x_i]$
and vanishing covariance
\begin{equation}
\mathbf{Cov}[x_i^k, x_j^l] \hspace*{2pt} \longrightarrow \hspace*{2pt} 0 \hspace*{5pt} \text{ for } \hspace*{5pt} d_G(x_i, x_j) \hspace*{2pt} \longrightarrow \hspace*{2pt} \infty
\end{equation}
for all $k, l \in \{1, 2\}$.
Furthermore, let $\hat{\bfx} := \bfx + \bfepsilon$ be another MRF, where $\bfepsilon = (\epsilon_1, ..., \epsilon_n)$ is an additive noise vector of i.i.d. variables
with $\mu_{\bfepsilon} := \mathbf{E}[\epsilon_i]$ and $\sigma^2_{\bfepsilon} := \mathbf{Var}[\epsilon_i]$.
Then the following holds true
\begin{equation}
\label{E_15041201}
\underset{n \hspace*{2pt}\rightarrow \hspace*{2pt} \infty}{\text{lim}} \mathbb{P}_{\bfx, \bfy}\left(\| \hat{\bfx} - \bfx \|_2 \hspace*{2pt}\leqslant \| \hat{\bfx} - \bfy \|_2 \right) = 1,
\end{equation}
provided
\begin{equation}
\label{E_19031029}
\mu_{\bfepsilon}^2 + \sigma^2_{\bfepsilon} \leqslant \frac{1}{2}\sigma^2.
\end{equation}
\end{itshape}
\end{theorem}
In the special example of the isotropic gaussian noise $\bfepsilon \sim \mathcal{N}(\boldsymbol{0}, \sigma^2_{\bfepsilon} \cdot I)$,
the condition in (\ref{E_19031029}) is reduced to $\sigma^2_{\bfepsilon} \leqslant \frac{1}{2}\sigma^2$.
Here, again $\bfx$ can be identified with the image of the conservative projection, which removes the specific type of noise from its arguments.
The equation in (\ref{E_15041201}) then implies its convergence to the orthogonal projection.

\section{Experiments}
\label{sec:section5}

\newcolumntype{g}{>{\columncolor{ColumnColor}}c}
\begin{table*}[t]
\scriptsize
  \caption{Experimental results for \textbf{anomaly segmentation} measured with \textbf{pixel-level AUROC} on the MVTec AD dataset.}
  \label{table_AUROC}
  \centering
  \begin{tabular}{ccccccccccccgg}
    \toprule
    Category & AnoGAN & VAE & LSR  & RIAD & CutPaste & InTra & DRAEM & SimpleNet & PatchCore & MSFlow & PNI & Model I & Model II\\
    \midrule
    carpet  &  54  &  78  & 94 & 96.3 & 98.3 & 99.2 & 95.5 & 98.2 & 98.7 & 99.4 & 99.4 & \textcolor{blue}{99.7} & \textcolor{red}{99.8}\\
    grid &       58  &  73  & 99 & 98.8  & 97.5 & 98.8 & \textcolor{blue}{99.7} & 98.8 & 98.8 & 99.4 & 99.2 & \textcolor{blue}{99.7} & \textcolor{red}{99.8}\\
    leather &  64 &  95  & 99 & 99.4 & 99.5 & 99.5 & 98.6 & 99.2 & 99.3 & \textcolor{red}{99.7} & \textcolor{blue}{99.6} & \textcolor{red}{99.7} & \textcolor{red}{99.7}\\
    tile &         50 &  80  & 88 & 89.1 &  90.5 & 94.4 & \textcolor{red}{99.2} & 97.0 & 96.3 & 98.2 & \textcolor{blue}{98.4} & \textcolor{red}{99.2} & \textcolor{red}{99.2}\\
    wood &     62  & 77  & 87 & 85.8 & 95.5 & 88.7 & 96.4 & 94.5 & 95.2 & \textcolor{blue}{97.1} & 97.0 & \textcolor{red}{98.4} & \textcolor{red}{98.4}\\
    \midrule
    avg. tex.  & 57.6 & 80.6 & 93.4 & 93.9 & 96.3 & 96.1 & 97.9 & 97.5 & 97.7 & 98.8 & 98.7 & \textcolor{blue}{99.3} & \textcolor{red}{99.4}\\
    \midrule
    bottle &  86  & 87 & 95 & 98.4 & 97.6 & 97.1 & \textcolor{red}{99.1} & 98.0 & 98.6 & \textcolor{blue}{99.0} & 98.9 & 98.6 & 98.9\\
    cable &  86  & 87 & 95 & 94.2 & 90.0 & 91.0 & 94.7 & 97.6 & \textcolor{blue}{98.7} & 98.5 & \textcolor{red}{99.1} & 98.2 & 98.5\\
    capsule &  84  & 74 & 93 & 92.8 & 97.4 & 97.7 & 94.3 & 98.9 & \textcolor{blue}{99.1} & \textcolor{blue}{99.1} & \textcolor{red}{99.3} & \textcolor{blue}{99.1} & \textcolor{blue}{99.1}\\
    hazelnut & 87  & 98 & 95 & 96.1 & 97.3 & 98.3 & \textcolor{red}{99.7} & 97.9 & 98.8 & 98.7 & \textcolor{blue}{99.4} & 98.9 & 99.1\\
    metal nut &  76  & 94 & 91 & 92.5 & 93.1 & 93.3 & \textcolor{red}{99.5} & 98.8 & 99.0 & \textcolor{blue}{99.3} & \textcolor{blue}{99.3} & 98.5 & 98.5\\
    pill & 87  & 83 & 91 & 95.7 & 95.7 & 98.3 & 97.6 & 98.6 & 98.6 & 98.8 & \textcolor{blue}{99.0} & \textcolor{red}{99.3} & \textcolor{red}{99.3}\\
    screw  & 80 & 97 & 96 & 98.8 & 96.7 & 99.5 & 97.6 & 99.3 & 99.5 & 99.1 & \textcolor{blue}{99.6} & \textcolor{red}{99.7} & \textcolor{red}{99.7}\\
    toothbrush & 90 & 94 & 97 & 98.9 & 98.1 & 98.9 & 98.1 & 98.5 & 98.9 & 98.5 & \textcolor{blue}{99.1} & \textcolor{blue}{99.1} & \textcolor{red}{99.4}\\
    transistor &  80 & 93 & 91 & 87.7 & 93.0 & 96.1 & 90.9 & 97.6 & 97.1 & 98.3 & 98.0 & \textcolor{blue}{98.6} & \textcolor{red}{98.9}\\
    zipper &  78 & 78 & 98 & 97.8 & 99.3 & 99.2 & 98.8 & 98.9 & 99.0 & 99.2 & 99.4 & \textcolor{blue}{99.5} & \textcolor{red}{99.6}\\
    \midrule
    avg. obj. & 83.4 & 88.5 & 94.6 & 95.3 & 95.8 & 96.9 & 97.0 & 98.4 & 98.7 & 98.8 & \textcolor{red}{99.1} & \textcolor{blue}{99.0} & \textcolor{red}{99.1} \\
    \midrule
    avg. all & 74.8 & 85.9 & 94.2 & 94.8 & 96.0 & 96.7 & 97.3 & 98.1 & 98.4 & 98.8 & 99.0 & \textcolor{blue}{99.1} & \textcolor{red}{99.2} \\
    \bottomrule
  \end{tabular}
\end{table*}

\begin{figure*}
\centering
\includegraphics[scale = 0.88]{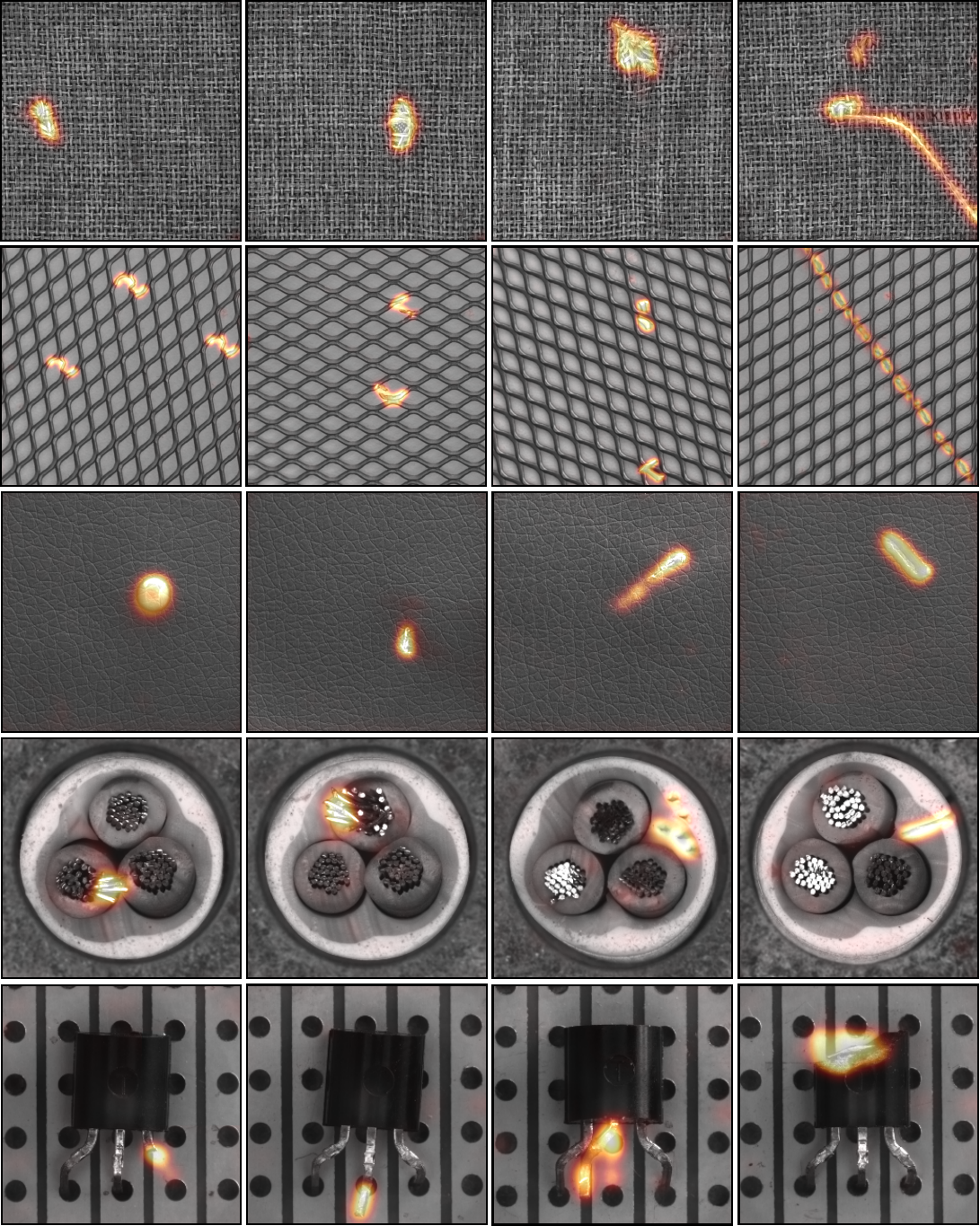}
\caption{
Illustration of our anomaly segmentation results as an overlay of the original image and the anomaly heatmap. Each row shows
four random examples from a category (carpet, grid, leather, cable and transistor) in the MVTec AD dataset.
}
\label{fig_bauer16}
\end{figure*}

\begin{table*}[t]
\scriptsize
  \caption{Experimental results for \textbf{anomaly recognition} measured with \textbf{image-level AUROC} on the MVTec AD dataset.}
  \label{table_image_AUROC}
  \centering
  \begin{tabular}{ccccccccccccgg}
    \toprule
    Category & AnoGAN & VAE  & LSR & RIAD & CutPaste & InTra & DRAEM & SimpleNet & PatchCore & MSFlow & PNI & Model I & Model II\\
    \midrule
    carpet  & 49 & 78 & 71 &  84.2 & 93.1 & 98.8 & 97.0 & \textcolor{blue}{99.7} & 98.2 & \textcolor{red}{100} & \textcolor{red}{100} & \textcolor{red}{100} & \textcolor{red}{100}\\
    grid      & 51 & 73 & 91 & 99.6 & \textcolor{blue}{99.9} & \textcolor{red}{100} & \textcolor{blue}{99.9} & 99.7 & 98.3 & 99.8 & 98.4 & \textcolor{red}{100} & \textcolor{red}{100}\\
    leather & 52 & 95 & \textcolor{blue}{96} & \textcolor{red}{100} & \textcolor{red}{100} & \textcolor{red}{100} & \textcolor{red}{100} & \textcolor{red}{100} & \textcolor{red}{100} & \textcolor{red}{100} & \textcolor{red}{100} & \textcolor{red}{100} & \textcolor{red}{100}\\
    tile        & 51 & 80 & 95 &  93.4 & 93.4 & 98.2 & 99.6 & \textcolor{blue}{99.8} & 98.9 & \textcolor{red}{100} & \textcolor{red}{100} & \textcolor{red}{100} & \textcolor{red}{100}\\
    wood    & 68 & 77 & 96 & 93.0 & 98.6 & 97.5 & 99.1 & \textcolor{red}{100} & \textcolor{blue}{99.9} & \textcolor{red}{100} & 99.6 & \textcolor{red}{100} & \textcolor{red}{100}\\
    \midrule
    avg. textures  & 54.2 & 80.6 & 89.8 & 94.0 & 97.0 & 98.9 & 99.1 & 99.8 & 99.0 & \textcolor{blue}{99.9} & 99.6 & \textcolor{red}{100} & \textcolor{red}{100}\\
    \midrule
    bottle         & 69 & 87 & 99 & \textcolor{blue}{99.9} & 98.3 & \textcolor{red}{100} & 99.2 & \textcolor{red}{100} & \textcolor{red}{100} & \textcolor{red}{100} & \textcolor{red}{100} & \textcolor{red}{100} & \textcolor{red}{100}\\
    cable         & 53 & 90 & 72 & 81.9 & 80.6 & 70.3 & 91.8 & \textcolor{red}{99.9} & 99.7 & 99.5 & \textcolor{blue}{99.8} & 98.0 & \textcolor{red}{99.9}\\
    capsule     & 58 & 74 & 68 & 88.4 & 96.2 & 86.5 & 98.5 & 97.7 & 98.1 & \textcolor{blue}{99.2} & \textcolor{red}{99.7} & 98.2 & 98.8\\
    hazelnut    & 50 & 98 & 94 & 83.3 & \textcolor{blue}{97.3} & 95.7 & \textcolor{red}{100} & \textcolor{red}{100} & \textcolor{red}{100} & \textcolor{red}{100} & \textcolor{red}{100} & \textcolor{red}{100} & \textcolor{red}{100}\\
    metal nut   & 50 & 94 & 83 & 88.5 & \textcolor{blue}{99.3} & 96.9 & 98.7 & \textcolor{red}{100} & \textcolor{red}{100} & \textcolor{red}{100} & \textcolor{red}{100} & \textcolor{red}{100} & \textcolor{red}{100}\\
    pill             & 62 & 83 & 68 & 83.8 & 92.4 & 90.2 & 98.9 & 99.0 & 97.1 & \textcolor{red}{99.6} & 96.9 & \textcolor{blue}{99.5} & \textcolor{red}{99.6}\\
    screw        & 35 & 97 & 80 & 84.5 & 86.3 & 95.7 & 93.9 & 98.2 & \textcolor{blue}{99.0} & 97.8 & \textcolor{red}{99.5} & 98.9 & 98.9\\
    toothbrush & 57 & 94 & 92 & \textcolor{red}{100} & 98.3 & \textcolor{red}{100} & \textcolor{red}{100} & \textcolor{blue}{99.7} & 98.9 & \textcolor{red}{100} & \textcolor{blue}{99.7} & \textcolor{red}{100} & \textcolor{red}{100}\\
    transistor   & 67 & 93 & 73 & 90.9 & 95.5 & 95.8 & 93.1 & \textcolor{red}{100} & \textcolor{blue}{99.7} & \textcolor{red}{100} & \textcolor{red}{100} & 98.8 & \textcolor{red}{100}\\
    zipper        & 59 & 78 & 97 & 98.1  & 99.4 & 99.4 & \textcolor{red}{100} & \textcolor{blue}{99.9} & 99.7 & \textcolor{red}{100} & \textcolor{blue}{99.9} & \textcolor{red}{100} & \textcolor{red}{100}\\
    \midrule
    avg. objects & 56.0 & 88.8 & 82.6 & 89.9 & 94.4 & 93.1 & 97.4 & 99.5 & 99.2 & \textcolor{blue}{99.6} & 99.5 & 99.3 & \textcolor{red}{99.7}\\
    \midrule
    avg. all        &  55.4 & 86.1 & 85.0 & 91.3 & 95.2 & 95.0 &  98.0  & 99.6 & 99.2 &  \textcolor{blue}{99.7} & 99.6 & 99.6 & \textcolor{red}{99.8}\\
    \bottomrule
  \end{tabular}
\end{table*}
In our experiments, we used the publicly available MVTec AD dataset \cite{BergmannFSS19, BergmannBFSS21}
-- a popular benchmark in the manufacturing domain,
which provides a substantial data variation including 5 texture and 10 object categories.
During training we artificially increase the amount of data by using simple data augmentation like rotation
and flipping depending on the category
while using $5\%$ of data as validation set.
For each category we trained a separate model
based on the proposed architectures (Model I and II).
For both models we used the same image resolution of $512 \times 512$ pixels for texture
and $256 \times 256$ pixels for object categories
and AdamOptimizer \cite{KingmaB14} with initial learning rate of $10^{-4}$.
We report the pixel-level and the image-level AUROC metric
to illustrate both segmentation and recognition performance in Table \ref{table_AUROC} and \ref{table_image_AUROC}, respectively.
We compare the results with a number of previous methods like
AnoGAN \cite{SchleglSWSL17}, VAE \cite{LiuLZKWBRC20}, LSR  \cite{LSR},
including top-ranking algorithms such as
RIAD \cite{ZavrtanikKS21}, CutPaste  \cite{LiSYP21}, InTra  \cite{InTra}, DRAEM \cite{ZavrtanikKS212}, SimpleNet \cite{LiuZXW23}, PatchCore \cite{RothPZSBG22}, MSFlow \cite{msflow-abs-2308-15300}, and PNI \cite{Jaehyeok2023}.
Our method achieves high performance with both models,
while Model II significantly outperforms Model I on the cable, capsule and transistor category
due to a reduction in false positive detections supporting our intuition around the CCM module.
To give a sense of visual quality of the resulting segmentation masks,
we show a few examples in Figure \ref{fig_bauer16}.
\section{Conclusion}
\label{sec:section6}
We focused on the important AD use case,
where the data distribution is supported by a lower-dimensional manifold.
On the example of imagery data, we introduced a self-supervised training framework,
which aims at learning a reconstruction model to repair artificially corrupted inputs based on a specific form of stochastic occlusions.
Altogether, the resulting training effect regularizes the model to produce locally consistent reconstructions
while replacing irregularities, therefore, acting as a filter that removes anomalous patterns.
We demonstrated the effectiveness of  our approach by achieving new state-of-the-art results on the MVTec AD dataset --
a challenging benchmark for visual anomaly detection in the manufacturing domain.

Additionally, we preformed a theoretical analysis of the proposed method
providing a few interesting insights.
As our main result, we showed that when the dimensionality of the inputs goes to infinity, a corresponding model
converges stochastically to the orthogonal projection of partially corrupted inputs onto the submanifold of uncorrupted examples.
In the case where the covariance between the input variables rapidly goes to zero with increasing distance,
the orthogonal projection maps its input in a way largely preserving the original content
supporting our intuition around the filtering behavior of the model.
Therefore, we can jointly perform detection and localization of corrupted regions by means of the pixel-wise reconstruction error.

Furthermore, we deepen the understanding of the relationship between the regularized autoencoders.
Specifically, we showed that the orthogonal projection provides an optimal solution for the RCAE,
which has been shown to be equivalent to the DAE for small variance of the corruption noise.
Our results extend the equivalence to more complex input modifications beyond the i.i.d. pixel corruptions,
which are not limited by the assumption of small variance.

\ifCLASSOPTIONcaptionsoff
  \newpage
\fi



\bibliographystyle{IEEEtran}
\bibliography{references}

\clearpage
\onecolumn 
\setcounter{section}{0}
\setcounter{page}{1}
\section*{Appendix}
\section{Auxiliary Statements}
In order to prove the formal statements in the main body of the paper, we first introduce a set of auxiliary statements
including Theorem \ref{theorem_WLBNMM} and Corollary \ref{cor_WLBNMM}, which themselves provide an additional contribution.
For convenience, we extend our notation as follows.
Given a vector $\bfx \in \mathbb{R}^n$, a set of indices $S \subset \{1, ..., n\} $ and its complement $\bar{S} := \{1, ..., n\} \setminus S$,
we interpret the restriction $\bfx_S$ in two different ways depending on the context either as an element $\bfx_S \in \mathbb{R}^{|S|}$ or
as an element $\bfx_S \in \mathbb{R}^{n}$ where the indices in $\bar{S}$ are set to zeros, that is, $\bfx_{\bar{S}} = {\bf0} \in \mathbb{R}^{|\bar{S}|}$, and vice versa.
Given such notation, we can write $\bfx = \bfx_S + \bfx_{\bar{S}}$.
\begin{lemma}
\label{l1}
Consider an $\bfx \in \mathbb{R}^n$, $n \in \mathbb{N}$ and $S \subset \{1, ..., n\} $, $\bar{S} := \{1, ..., n\} \setminus S$.
 The following holds true for all $p \in \mathbb{N}_+:$
\begin{equation}
\|\bfx_S \pm \bfx_{\bar{S}}\|^p_p = \|\bfx_S\|^p_p \pm \|\bfx_{\bar{S}}\|^p_p.
\end{equation}
\end{lemma}
The proof follows directly by writing out: $\|\bfx_S \pm \bfx_{\bar{S}}\|^p_p = \left(\sum_{i \in S} (x_i \pm 0)^p \pm \sum_{j \in \bar{S}} (0 \pm x_j)^p \right)^{\frac{1}{p} \cdot p} = \|\bfx_S\|^p_p + \|\bfx_{\bar{S}}\|^p_p$.

\begin{lemma}
\label{l3}
For all $\hat{\bfx}, \bfx, \bfy \in \mathbb{R}^n, n \in \mathbb{N}$ with $\hat{\bfx}_{\bar{S}} = \bfx_{\bar{S}}$ for some non-empty sets $S \subset \{1, ..., n\} $, $\bar{S} := \{1, ..., n\} \setminus S$,
the following holds true for all $p \in \mathbb{N}_+:$
\begin{equation}
\| \hat{\bfx} - \bfx \|_p \leqslant \| \hat{\bfx} - \bfy \|_p  \Longleftrightarrow \| \hat{\bfx}_S - \bfx_S \|^p_p - \| \hat{\bfx}_S - \bfy_S \|^p_p \leqslant \| \bfx_{\bar{S}} - \bfy_{\bar{S}} \|^p_p.
\end{equation}
\end{lemma}
\begin{proof}
\begin{equation*}
\begin{aligned}
 \| \hat{\bfx} - \bfx \|_p \leqslant \| \hat{\bfx} - \bfy \|_p &\Longleftrightarrow \| \hat{\bfx} - \bfx \|^p_p \leqslant \| \hat{\bfx} - \bfy \|^p_p \\
 &\Longleftrightarrow \| (\hat{\bfx}_S - \bfx_S) + (\hat{\bfx}_{\bar{S}} - \bfx_{\bar{S}}) \|^p_p \leqslant \| (\hat{\bfx}_S - \bfy_S) + (\hat{\bfx}_{\bar{S}} - \bfy_{\bar{S}}) \|^p_p \\
 &\overset{(*)}{\Longleftrightarrow} \| \hat{\bfx}_S - \bfx_S \|^p_p \leqslant \| \hat{\bfx}_S - \bfy_S \|^p_p + \| \hat{\bfx}_{\bar{S}} - \bfy_{\bar{S}} \|^p_p \\
 & \Longleftrightarrow \| \hat{\bfx}_S - \bfx_S \|^p_p - \| \hat{\bfx}_S - \bfy_S \|^p_p \leqslant \| \hat{\bfx}_{\bar{S}} - \bfy_{\bar{S}} \|^p_p
\end{aligned}
\end{equation*}
where in step $(*)$ we used Lemma \ref{l1} and our assumption our assumption $\hat{\bfx}_{\bar{S}} = \bfx_{\bar{S}}$.
\end{proof}

\section{Weak Law of Large Numbers for MRFs.}
Here we derive a version of the weak law of large numbers adjusted to our case of MRFs with identically distributed but non-independent variables.
\begin{theorem}[Weak Law of Large Numbers for MRFs]
\label{theorem_WLBNMM}
Let $(\{x_1, ..., x_n\})_{n \in \mathbb{N}}$ be a sequence of MRFs
over identically distributed variables $x_i$ with finite variance $\mathbf{Var}[x_i] < \infty$
and vanishing covariance
\begin{equation}
\mathbf{Cov}[x_i, x_j] \hspace*{2pt} \longrightarrow \hspace*{2pt} 0 \hspace*{5pt} \text{ for } \hspace*{5pt} d_G(x_i, x_j) \hspace*{2pt} \longrightarrow \hspace*{2pt} \infty.
\end{equation}
Then the following is true:
\begin{equation}
\frac{1}{n} \sum_{i=1}^n x_i \hspace*{5pt}\overset{\mathbf{P}}{\longrightarrow} \hspace*{5pt} \mathbf{E}[x_1] \hspace*{10pt} \text{for} \hspace*{10pt} n \hspace*{5pt} {\longrightarrow} \hspace*{5pt} \infty.
\end{equation}
\end{theorem}

\begin{proof}
First we reorganize the individual terms in the definition of variance as follows:
\begin{equation*}
\begin{aligned}
\mathbf{Var}\left[\sum_{i=1}^n x_i \right]  =& \hspace*{3pt} \mathbf{E}\left[\left(\sum_{i=1}^n x_i - \mathbf{E}\left[ \sum_{i=1}^n x_i\right]\right)^2 \right]\\
 =& \hspace*{3pt} \mathbf{E}\left[\left(\sum_{i=1}^n x_i - \mathbf{E}\left[ x_i\right] \right)^2 \right]\\
 =& \hspace*{3pt} \sum_{i=1}^n \sum_{j=1}^n \mathbf{Cov}[x_i, x_j]\\
  =& \hspace*{3pt} \sum_{i=1}^n \mathbf{Var}[x_i] + 2 \sum_{i=1}^n \sum_{j=i+1}^n \mathbf{Cov}[x_i, x_j].
\end{aligned}
\end{equation*}
Let $\epsilon > 0$. Since $\mathbf{Cov}[x_i, x_j] \hspace*{2pt} \longrightarrow \hspace*{2pt} 0$ for $d_G(x_i, x_j) \hspace*{2pt} \longrightarrow \hspace*{2pt} \infty$,
there is an $L \in \mathbb{N}$ such that for all $i, j$ satisfying $d_G(x_i, x_j) \hspace*{3pt}> L$ we get $\left|\mathbf{Cov}[x_i,x_j]\right| < \epsilon$.
Later we will consider $n  \longrightarrow \infty$. Therefore, we can assume $n > L$ and
split the inner sum of the covariance term in the last equation in two sums as follows.
Here, we abbreviate the index set $\{j \in \mathbb{N} \colon i+1 \leqslant j \leqslant n, d_G(x_i, x_j) \leqslant L\}$ and $\{j \in \mathbb{N} \colon i+1 \leqslant j \leqslant n, d_G(x_i, x_j) > L\}$
by $d_G(x_i, x_j) \leqslant L$ and  $d_G(x_i, x_j) > L$, respectively.
That is,
\begin{equation*}
\sum_{i=1}^n \sum_{j=i+1}^n \mathbf{Cov}[x_i, x_j] = \sum_{i=1}^n \left( \sum_{j \colon d_G(x_i, x_j) \leqslant L} \underbrace{\mathbf{Cov}[x_i, x_j]}_{\leqslant \sigma^2} + \sum_{j \colon d_G(x_i, x_j) > L}  \underbrace{\mathbf{Cov}[x_i, x_j]}_{< \epsilon} \right) < n \cdot L \cdot \sigma^2 + n^2 \cdot \epsilon,
\end{equation*}
where in the last step we used, on the one hand,
the Cauchy-Schwarz inequality $|\mathbf{Cov}[x_i, x_j]| \leqslant \sqrt{\mathbf{Var}[x_i] \cdot \mathbf{Var}[x_j]} = \mathbf{Var}[x_i] =: \sigma^2$,
and on the other hand, an upper bound $L$ on the cardinality of the running index set in the second sum.
Altogether, we get the following estimation:
\begin{equation*}
\begin{aligned}
\frac{1}{n^2}\mathbf{Var}\left[\sum_{i=1}^n x_i \right]  <& \hspace*{3pt} \frac{\sigma^2}{n} + \frac{2 \cdot L \cdot \sigma^2}{n} + 2 \cdot \epsilon.
\end{aligned}
\end{equation*}
Since the first two terms on the right hand side converge to zero (for $n \longrightarrow \infty$) and $\epsilon$ is chosen arbitrarily small,
this implies:
\begin{equation}
\label{eq_Var_zero}
\underset{n \hspace*{2pt}\rightarrow \hspace*{2pt} \infty}{\text{lim}} \frac{1}{n^2} \mathbf{Var}\left[\sum_{i=1}^n x_i \right] = 0.
\end{equation}
\noindent Finally, for all $\epsilon > 0$:
\begin{equation*}
\mathbf{P}\left(\left| \frac{1}{n} \sum_{i=1}^n x_i - \mathbf{E}[x_1] \right| > \epsilon \right) \hspace*{5pt} \overset{(*)}{\leqslant} \hspace*{5pt} \frac{1}{\epsilon^2}\mathbf{Var}\left[\frac{1}{n} \sum_{i=1}^n x_i \right]
= \hspace*{5pt} \frac{1}{\epsilon^2 n^2}\mathbf{Var}\left[\sum_{i=1}^n x_i \right]\\
\end{equation*}
where in step $(*)$ we used the Chebyshev's inequality.
The convergence
\begin{equation}
\underset{n \hspace*{2pt}\rightarrow \hspace*{2pt} \infty}{\text{lim}} \mathbf{P}\left(\left| \frac{1}{n} \sum_{i=1}^n x_i - \mathbf{E}[x_1] \right| > \epsilon \right) = 0
\end{equation}
follows from equation (\ref{eq_Var_zero}).
\end{proof}

\section{Corollary of Theorem \ref{theorem_WLBNMM}}
Applying Theorem \ref{theorem_WLBNMM} to our discussion in the paper
we get the following useful corollary.
\begin{corollary}
\label{cor_WLBNMM}
Consider a pair $\bfx, \bfy$ of independent MRFs
over identically distributed variables $x_i, y_j$ with finite fourth moment $\mathbf{E}[x_i^4] < \infty$
and vanishing covariance according to
\begin{equation}
\label{cor_WLBNMM_0}
\mathbf{Cov}[x_i^k, x_j^l] \hspace*{2pt} \longrightarrow \hspace*{2pt} 0 \hspace*{5pt} \text{ for } \hspace*{5pt} d_G(x_i, x_j) \hspace*{2pt} \longrightarrow \hspace*{2pt} \infty
\end{equation}
for all $k, l \in \{1, 2\}$.
The following statements are true:
\begin{align}
\label{eq_cor_WLBNMM_1}
(a)&\hspace*{10pt} \frac{1}{|S|} \|\bfx_S\|^2_2 \hspace*{5pt}\overset{\mathbf{P}}{\longrightarrow} \hspace*{5pt} \mu^2 + \sigma^2 &\text{for} \hspace*{10pt} |S| \hspace*{5pt} {\longrightarrow} \hspace*{5pt} \infty,&\\
\label{eq_cor_WLBNMM_2}
(b)&\hspace*{10pt} \frac{1}{|S|} \|\bfx_S - \bfy_S \|^2_2 \hspace*{5pt}\overset{\mathbf{P}}{\longrightarrow} \hspace*{5pt} 2\sigma^2 &\text{for} \hspace*{10pt} |S| \hspace*{5pt} {\longrightarrow} \hspace*{5pt} \infty,&\\
\label{eq_cor_WLBNMM_3}
(c)&\hspace*{10pt} \frac{1}{|S|} \bfx_S^{\top} \bfy_S \hspace*{5pt}\overset{\mathbf{P}}{\longrightarrow} \hspace*{5pt} \mu^2 &\text{for} \hspace*{10pt} |S| \hspace*{5pt} {\longrightarrow} \hspace*{5pt} \infty,& \\
\label{eq_cor_WLBNMM_4}
(d)&\hspace*{10pt} \cos \phi[\bfx_S, \bfy_S] \hspace*{5pt}\overset{\mathbf{P}}{\longrightarrow} \hspace*{5pt} \frac{\mu^2}{\mu^2 + \sigma^2} &\text{for} \hspace*{10pt} |S| \hspace*{5pt} {\longrightarrow} \hspace*{5pt} \infty,&
\end{align}
where  $\mu := \mathbf{E}[x_i]$, $\sigma^2 := \mathbf{Var}[x_i]$, $\phi[\bfx_S, \bfy_S]$ is the angle between the two vectors and 
$S \subseteq \mathbb{N}$ denotes a subset of the variable indices.
\end{corollary}

\begin{proof}
First we prove the statement $(a)$ in (\ref{eq_cor_WLBNMM_1}).
By setting $z_i := x_i^2$ we get:
\begin{equation}
\mathbf{E}[z_1] = \mathbf{E}[x_1^2] = \mathbf{Var}[x_1] + \mathbf{E}[x_1]^2 = \sigma^2 + \mu^2.
\end{equation}
On the other hand, we get:
\begin{equation}
\frac{1}{|S|}\|\bfx_S \|^2 = \frac{1}{|S|} \sum_{i \in S} x_i^2 = \frac{1}{|S|} \sum_{i \in S} z_i.
\end{equation}
Since we assumed $\mathbf{E}[x_i^4] < \infty$, it holds
\begin{equation*}
\mathbf{Var}[z_i] = \mathbf{Var}[x_i^2] = \mathbf{E}[x_i^4] - \mathbf{E}[x_i^2]^2 = \mathbf{E}[x_i^4] - (\sigma^2 + \mu^2)^2 < \infty.
\end{equation*}
Furthermore, due to our assumption in (\ref{cor_WLBNMM_0}):
\begin{equation*}
\mathbf{Cov}[z_i, z_j] = \mathbf{Cov}[x_i^2, x_j^2] \hspace*{2pt} \longrightarrow \hspace*{2pt} 0 \hspace*{5pt} \text{ for } \hspace*{5pt} d_G(x_i, x_j) \hspace*{2pt} \longrightarrow \hspace*{2pt} \infty.
\end{equation*}
That is, the variables $z_i$ satisfy all the requirements in Theorem \ref{theorem_WLBNMM}.
Applying this theorem directly to the above derivations proves the claim in (\ref{eq_cor_WLBNMM_1}).

Now we prove the statement $(b)$ in (\ref{eq_cor_WLBNMM_2}).
For this purpose we set $z_i := (x_i - y_i)^2$.
It holds:
\begin{equation*}
\mathbf{E}[z_1]  = \mathbf{E}[(x_1 - y_1)^2] = \mathbf{E}[x_1^2] - 2 \mathbf{E}[x_1 y_1] + \mathbf{E}[y_1^2]
=  2 \left( \mathbf{E}[x_1^2] - \mathbf{E}[x_1]^2 \right)
=  2 \mathbf{Var}[x_1]
=  2 \sigma^2.
\end{equation*}
On the other hand we get:
\begin{equation}
\frac{1}{|S|}\|\bfx_S - \bfy_S \|^2 = \frac{1}{|S|} \sum_{i \in S} (x_i - y_i)^2 = \frac{1}{|S|} \sum_{i \in S} z_i.
\end{equation}
Things remaining to be shown are that the corresponding variance is finite and that the covariance vanishes.
It holds:
\begin{equation*}
\scriptsize
\begin{aligned}
& \hspace*{3pt} \mathbf{E}[(x_i - y_i)^2 (x_j - y_j)^2] \\
 =& \hspace*{3pt} \mathbf{E}[(x_i^2 - 2 x_i y_i + y_i^2) (x_j ^2 - 2 x_j y_j + y_j^2)] \\
 =& \hspace*{3pt} \mathbf{E}[x_i^2 x_j^2 - 2 x_i^2 x_j y_j + x_i^2 y_j^2 - 2 x_i y_i x_j^2 + 4 x_i y_i x_j y_j - 2 x_i y_i y_j^2 + y_i^2 x_j^2 - 2 y_i^2 x_j y_j + y_i^2 y_j^2] \\
 =& \hspace*{3pt} \mathbf{E}[x_i^2 x_j^2] - 2 \mathbf{E}[x_i^2 x_j y_j] + \mathbf{E}[x_i^2 y_j^2] - 2 \mathbf{E}[x_i y_i x_j^2 ]+ 4 \mathbf{E}[x_i y_i x_j y_j] - 2 \mathbf{E}[x_i y_i y_j^2] + \mathbf{E}[y_i^2 x_j^2] - 2 \mathbf{E}[y_i^2 x_j y_j] + \mathbf{E}[y_i^2 y_j^2]\\
 =& \hspace*{3pt} \mathbf{E}[x_i^2 x_j^2] - \blueOn 2 \mathbf{E}[x_i^2 x_j]\mathbf{E}[y_j] \blackOn + \redOn \mathbf{E}[x_i^2] \mathbf{E}[y_j^2] \blackOn - \blueOn 2 \mathbf{E}[x_i x_j^2] \mathbf{E}[y_i ] \blackOn + \brownOn 4 \mathbf{E}[x_i x_j] \mathbf{E}[y_i y_j] \blackOn - \blueOn 2 \mathbf{E}[x_i] \mathbf{E}[y_i y_j^2] \blackOn + \redOn \mathbf{E}[y_i^2] \mathbf{E}[x_j^2] \blackOn - \blueOn 2 \mathbf{E}[x_j]\mathbf{E}[y_i^2 y_j] \blackOn + \mathbf{E}[y_i^2 y_j^2]\\
 =& \hspace*{3pt} 2 \mathbf{E}[x_i^2 x_j^2] - \blueOn 8 \mathbf{E}[x_i^2 x_j]\mathbf{E}[x_j] \blackOn + \redOn 2 \mathbf{E}[x_i^2] \mathbf{E}[x_j^2] \blackOn + \brownOn 4 \mathbf{E}[x_i x_j]^2 \blackOn .
\end{aligned}
\end{equation*}
On the other hand:
\begin{equation*}
\scriptsize
\begin{aligned}
& \hspace*{3pt} \mathbf{E}[(x_i - y_i)^2] \mathbf{E}[(x_j - y_j)^2] \\
 =& \hspace*{3pt} \mathbf{E}[(x_i^2 - 2 x_i y_i + y_i^2)] \mathbf{E}[(x_j ^2 - 2 x_j y_j + y_j^2)] \\
 =& \hspace*{3pt} (\mathbf{E}[x_i^2] - 2 \mathbf{E}[x_i y_i] + \mathbf{E}[y_i^2]) (\mathbf{E}[x_j^2] - 2 \mathbf{E}[x_j y_j] + \mathbf{E}[y_j^2]) \\
 =& \hspace*{3pt} \mathbf{E}[x_i^2] \mathbf{E}[x_j^2] - 2 \mathbf{E}[x_i^2] \mathbf{E}[x_j y_j] + \mathbf{E}[x_i^2] \mathbf{E}[y_j^2] - 2 \mathbf{E}[x_i y_i] \mathbf{E}[x_j^2] + 4 \mathbf{E}[x_i y_i] \mathbf{E}[x_j y_j] - 2 \mathbf{E}[x_i y_i] \mathbf{E}[y_j^2] + \mathbf{E}[y_i^2] \mathbf{E}[x_j^2] \\
  & \hspace*{1pt} - 2 \mathbf{E}[y_i^2] \mathbf{E}[x_j y_j] + \mathbf{E}[y_i^2] \mathbf{E}[y_j^2] \\
 =& \hspace*{3pt} \mathbf{E}[x_i^2] \mathbf{E}[x_j^2] - 2 \mathbf{E}[x_i^2] \mathbf{E}[x_j] \mathbf{E}[y_j] + \mathbf{E}[x_i^2] \mathbf{E}[y_j^2] - 2 \mathbf{E}[x_i] \mathbf{E}[y_i] \mathbf{E}[x_j^2] + 4 \mathbf{E}[x_i] \mathbf{E}[ y_i] \mathbf{E}[x_j] \mathbf{E}[y_j] - 2 \mathbf{E}[x_i] \mathbf{E}[y_i] \mathbf{E}[y_j^2] + \mathbf{E}[y_i^2] \mathbf{E}[x_j^2]\\
 & \hspace*{1pt} - 2 \mathbf{E}[y_i^2] \mathbf{E}[x_j] \mathbf{E}[y_j] + \mathbf{E}[y_i^2] \mathbf{E}[y_j^2] \\
 =& \hspace*{3pt} \mathbf{E}[x_i^2] \mathbf{E}[x_j^2] - \blueOn 2 \mathbf{E}[x_i^2] \mathbf{E}[x_j]^2 \blackOn + \mathbf{E}[x_i^2] \mathbf{E}[x_j^2] - \blueOn 2 \mathbf{E}[x_i]^2 \mathbf{E}[x_j^2] \blackOn + \redOn 4 \mathbf{E}[x_i]^2 \mathbf{E}[x_j]^2 \blackOn - \blueOn 2 \mathbf{E}[x_i]^2 \mathbf{E}[x_j^2] \blackOn + \mathbf{E}[x_i^2] \mathbf{E}[x_j^2] - \blueOn 2 \mathbf{E}[x_i^2] \mathbf{E}[x_j]^2 \blackOn + \mathbf{E}[x_i^2] \mathbf{E}[x_j^2] \\
  =& \hspace*{3pt} 4 \mathbf{E}[x_i^2] \mathbf{E}[x_j^2] - \blueOn 8 \mathbf{E}[x_i^2] \mathbf{E}[x_j]^2 \blackOn + \redOn 4 \mathbf{E}[x_i]^2 \mathbf{E}[x_j]^2 \blackOn.
\end{aligned}
\end{equation*}
Based on the above derivations we now analyze the covariance term
\begin{equation*}
\mathbf{Cov}[z_i, z_j] = \mathbf{E}[(x_i - y_i)^2 (x_j - y_j)^2] - \mathbf{E}[(x_i - y_i)^2] \mathbf{E}[(x_j - y_j)^2]
\end{equation*}
in the limit $d_G(z_i, z_j) \hspace*{2pt} \longrightarrow \hspace*{2pt} \infty$.
Because of our assumption in (\ref{cor_WLBNMM_0}),
the terms $\mathbf{E}[x_i^2 x_j^2]$, $\mathbf{E}[x_i^2 x_j]$ and $\mathbf{E}[x_i x_j]$ converge
against $\mathbf{E}[x_i^2] \mathbf{E}[x_j^2]$, $\mathbf{E}[x_i^2] \mathbf{E}[x_j]$ and $\mathbf{E}[x_i] \mathbf{E}[x_j]$, respectively.
It follows:
\begin{equation*}
\begin{aligned}
\underset{d_G(z_i, z_j) \hspace*{2pt}\rightarrow \hspace*{2pt} \infty}{\text{lim}} \mathbf{E}[(x_i - y_i)^2 (x_j - y_j)^2]
&= \underset{d_G(z_i, z_j) \hspace*{2pt}\rightarrow \hspace*{2pt} \infty}{\text{lim}} \left( 2 \mathbf{E}[x_i^2 x_j^2] - 8 \mathbf{E}[x_i^2 x_j]\mathbf{E}[x_j] + 2 \mathbf{E}[x_i^2] \mathbf{E}[x_j^2] + 4 \mathbf{E}[x_i x_j]^2 \right)\\
&= 2 \mathbf{E}[x_i^2] \mathbf{E}[x_j^2] - 8 \mathbf{E}[x_i^2] \mathbf{E}[x_j] \mathbf{E}[x_j] + 2 \mathbf{E}[x_i^2] \mathbf{E}[x_j^2] + 4 \mathbf{E}[x_i]^2 \mathbf{E}[x_j]^2 \\
&= 4 \mathbf{E}[x_i^2] \mathbf{E}[x_j^2] - 8 \mathbf{E}[x_i^2] \mathbf{E}[x_j]^2 + 4 \mathbf{E}[x_i]^2 \mathbf{E}[x_j]^2 \\
&= \mathbf{E}[(x_i - y_i)^2] \mathbf{E}[(x_j - y_j)^2].
\end{aligned}
\end{equation*}
Therefore, $\mathbf{Cov}[z_i, z_j] \hspace*{2pt} \longrightarrow \hspace*{2pt} 0 \hspace*{5pt} \text{ for } \hspace*{5pt} d_G(z_i, z_j) \hspace*{2pt} \longrightarrow \hspace*{2pt} \infty$.
Finally, by writing out
\begin{equation*}
\mathbf{Var}[z_i] = \mathbf{E}[z_i^2] - \mathbf{E}[z_i]^2 = \mathbf{E}[(x_i - y_i)^4] - \mathbf{E}[(x_i - y_i)^2]^2
\end{equation*}
and using the independence assumption of $x_i$ and $y_i$,
we can see that all monomials has the degree at most $4$. Due to our assumption of finite fourth moment,
this implies $\mathbf{Var}[z_i] < \infty$.
Altogether, the variables $z_i$ satisfy the requirements in Theorem \ref{theorem_WLBNMM}.
Applying this theorem directly to the above derivations proves the claim in (\ref{eq_cor_WLBNMM_2}).

The statement $(c)$ in (\ref{eq_cor_WLBNMM_3}) can be proven in a similar way.
Alternatively, we can use the statements in $(a)$ and $(b)$ and apply the limit algebra of convergence in probability
\begin{equation*}
\underbrace{\frac{1}{|S|} \|\bfx_S - \bfy_S\|^2}_{\overset{\mathbf{P}}{\longrightarrow} \hspace*{5pt} 2\sigma^2} = \underbrace{\frac{1}{|S|}\|\bfx_S\|^2}_{\overset{\mathbf{P}}{\longrightarrow} \hspace*{5pt} \mu^2 + \sigma^2} - \frac{2}{|S|}\bfx_S^{\top}\bfy_S + \underbrace{\frac{1}{|S|}\|\bfy_S\|^2}_{\overset{\mathbf{P}}{\longrightarrow} \hspace*{5pt} \mu^2 + \sigma^2},
\end{equation*}
which implies $\frac{1}{|S|} \bfx_S^{\top} \bfy_S \hspace*{5pt}\overset{\mathbf{P}}{\longrightarrow} \hspace*{5pt} \mu^2$.

Similarly, we prove the statement $(d)$ in (\ref{eq_cor_WLBNMM_4}) as follows:
\begin{equation*}
\begin{aligned}
& \hspace*{3pt} \cos \phi[\bfx_S, \bfy_S] \cdot \|\bfx_S\| \cdot \|\bfy_S\| =  \bfx_S^{\top} \bfy_S \\
\Longleftrightarrow& \hspace*{3pt} \cos \phi[\bfx_S, \bfy_S] \cdot \underbrace{\frac{1}{\sqrt{|S|}}\|\bfx_S\|}_{\overset{\mathbf{P}}{\longrightarrow} \hspace*{5pt} \sqrt{\mu^2 + \sigma^2}} \cdot \underbrace{\frac{1}{\sqrt{|S|}}\|\bfy_S\|}_{\overset{\mathbf{P}}{\longrightarrow} \hspace*{5pt} \sqrt{\mu^2 + \sigma^2}} =  \underbrace{\frac{1}{|S|} \bfx_S^{\top} \bfy_S}_{\overset{\mathbf{P}}{\longrightarrow} \hspace*{5pt} \mu^2} \\
\Longrightarrow& \hspace*{3pt} \cos \phi[\bfx_S, \bfy_S] \hspace*{5pt}\overset{\mathbf{P}}{\longrightarrow} \hspace*{5pt} \frac{\mu^2}{\mu^2 + \sigma^2} \hspace*{5pt} \text{ for }\hspace*{5pt} |S| \hspace*{5pt} {\longrightarrow} \hspace*{5pt} \infty.
\end{aligned}
\end{equation*}
\end{proof}

\section{Proof of Theorem \ref{theorem_1}}
\begin{proof}
Consider the case:
\begin{equation}
\|\bfx_{\bar{S}} - \bfy_{\bar{S}} \|^2_2 \hspace*{3pt}< \|\hat{\bfx}_S - \bfx_S\|^2_2 - \|\hat{\bfx}_S - \bfy_S\|^2_2 \hspace*{2pt}\leqslant \|\hat{\bfx}_S - \bfx_S\|^2_2 \leqslant |S|,
\end{equation}
where the last inequality holds due to $\hat{x}_i, y_j \in [0, 1]$. This implies:
\begin{equation}
\label{ineq_1}
\{(\bfx, \bfy) \colon \|\bfx_{\bar{S}} - \bfy_{\bar{S}} \|^2 \hspace*{3pt}< \|\hat{\bfx}_S - \bfx_S\|^2 - \|\hat{\bfx}_S - \bfy_S\|^2\} \subseteq \{(\bfx, \bfy) \colon \|\bfx_{\bar{S}} - \bfy_{\bar{S}} \|^2 <  |S|\}
\end{equation}
Consider now the following derivation:
\begin{equation*}
\begin{aligned}
\|\bfx_{\bar{S}} - \bfy_{\bar{S}} \|^2 \hspace*{2pt}<  |S| \Longleftrightarrow& \hspace*{3pt} \frac{1}{|\bar{S}|}\|\bfx_{\bar{S}} - \bfy_{\bar{S}} \|^2 \hspace*{2pt}<  \frac{|S|}{|\bar{S}|}\\
\Longrightarrow& \hspace*{3pt} \frac{1}{|\bar{S}|}\|\bfx_{\bar{S}} - \bfy_{\bar{S}} \|^2 \hspace*{2pt}<  2 \sigma^2\\
\Longrightarrow& \hspace*{3pt} \left| \frac{1}{|\bar{S}|}\|\bfx_{\bar{S}} - \bfy_{\bar{S}} \|^2 \hspace*{2pt} - 2 \sigma^2 \right| > 0,
\end{aligned}
\end{equation*}
where in the second step we used $|S| \hspace*{2pt}\leqslant 2 \sigma^2 |\bar{S}|$.
Therefore, it follows from (\ref{ineq_1}):
\begin{equation*}
\begin{aligned}
& \hspace*{3pt} \{(\bfx, \bfy) \colon \|\bfx_{\bar{S}} - \bfy_{\bar{S}} \|^2 \hspace*{3pt}< \|\hat{\bfx}_S - \bfx_S\|^2 - \|\hat{\bfx}_S - \bfy_S\|^2\}\\
\subseteq& \hspace*{3pt} \{(\bfx, \bfy) \colon \left| \frac{1}{|\bar{S}|}\|\bfx_{\bar{S}} - \bfy_{\bar{S}} \|^2 \hspace*{2pt} - 2 \sigma^2 \right| > 0\}\\
=& \hspace*{3pt} \{(\bfx, \bfy) \colon \left| \frac{1}{|\bar{S}|}\|\bfx_{\bar{S}} - \bfy_{\bar{S}} \|^2 \hspace*{2pt} - 2 \sigma^2 \right| > 1\} \hspace*{3pt} \cup \bigcup_{k \geqslant 2}\{(\bfx, \bfy) \colon \frac{1}{k-1} \geqslant \left|\frac{1}{|\bar{S}|}\|\bfx_{\bar{S}} - \bfy_{\bar{S}} \|^2 \hspace*{2pt} - 2 \sigma^2 \right| > \frac{1}{k}\}
\end{aligned}
\end{equation*}
Using the above derivations we now upper bound the probability of the corresponding events,
where $\mathbf{P} = \mathbf{P}_{\bfx, \bfy}$ denotes a joint probability distribution over $(\bfx, \bfy)$:
\begin{equation*}
\begin{aligned}
\mathbf{P}\left(\|\bfx_{\bar{S}} - \bfy_{\bar{S}} \|^2 \hspace*{3pt}< \|\hat{\bfx}_S - \bfx_S\|^2 - \|\hat{\bfx}_S - \bfy_S\|^2\right)
&\leqslant \mathbf{P}\left( \bigcup_{k}\{ \frac{1}{k-1} \geqslant \left|\frac{1}{|\bar{S}|}\|\bfx_{\bar{S}} - \bfy_{\bar{S}} \|^2 \hspace*{2pt} - 2 \sigma^2 \right| > \frac{1}{k}\} \right)\\
&\overset{(*)}{=} \sum_{k} \mathbf{P}\left( \frac{1}{k-1} \geqslant \left|\frac{1}{|\bar{S}|}\|\bfx_{\bar{S}} - \bfy_{\bar{S}} \|^2 \hspace*{2pt} - 2 \sigma^2 \right| > \frac{1}{k} \right),
\end{aligned}
\end{equation*}
where in step $(*)$ we used the fact that all the sets in the union are disjoint.
Note that we use a short notation $\bigcup_k$ and $\sum_k$ which runs over all previously defined terms (including $k = 1$).
Next we consider the above derivations in the limit $n \longrightarrow \infty$ for $S \subseteq \{1, ..., n\}$, $\bar{S} := \{1, ..., n\} \setminus S$
with a fixed ratio $|S| / |\bar{S}| \hspace*{2pt}\leqslant 2 \sigma^2$.
It holds:
\begin{equation*}
\begin{aligned}
\underset{n \hspace*{2pt}\rightarrow \hspace*{2pt} \infty}{\text{lim}}  \mathbf{P}\left(\|\bfx_{\bar{S}} - \bfy_{\bar{S}} \|^2 \hspace*{3pt}< \|\hat{\bfx}_S - \bfx_S\|^2 - \|\hat{\bfx}_S - \bfy_S\|^2\right)
&\leqslant \underset{n \hspace*{2pt}\rightarrow \hspace*{2pt} \infty}{\text{lim}} \underbrace{\sum_{k} \mathbf{P}\left( \frac{1}{k-1} \geqslant \left|\frac{1}{|\bar{S}|}\|\bfx_{\bar{S}} - \bfy_{\bar{S}} \|^2 \hspace*{2pt} - 2 \sigma^2 \right| > \frac{1}{k} \right)}_{\leqslant 1}\\
&\overset{(*)}{=} \sum_{k} \underset{n \hspace*{2pt}\rightarrow \hspace*{2pt} \infty}{\text{lim}} \mathbf{P}\left( \frac{1}{k-1} \geqslant \left|\frac{1}{|\bar{S}|}\|\bfx_{\bar{S}} - \bfy_{\bar{S}} \|^2 \hspace*{2pt} - 2 \sigma^2 \right| > \frac{1}{k} \right)\\
&\leqslant \sum_{k} \underbrace{\underset{n \hspace*{2pt}\rightarrow \hspace*{2pt} \infty}{\text{lim}} \mathbf{P}\left( \left|\frac{1}{|\bar{S}|}\|\bfx_{\bar{S}} - \bfy_{\bar{S}} \|^2 \hspace*{2pt} - 2 \sigma^2 \right| > \frac{1}{k} \right)}_{= 0}\\
&= 0,
\end{aligned}
\end{equation*}
where in step $(*)$ we can swap the limit and the sum signs because the series converges for all $n \in \mathbb{N}$.
The last step follows from Corollary \ref{cor_WLBNMM}.

To summarize, we've just shown:
\begin{equation}
\label{e_t2_for_t3}
\underset{n \hspace*{2pt}\rightarrow \hspace*{2pt} \infty}{\text{lim}}  \mathbf{P}_{\bfx, \bfy}\left(\|\hat{\bfx}_S - \bfx_S\|^2 - \|\hat{\bfx}_S - \bfy_S\|^2 \leqslant \|\bfx_{\bar{S}} - \bfy_{\bar{S}}\|^2\right) = 1.
\end{equation}
According to Lemma \ref{l3} it holds:
\begin{equation}
\label{E_2024_02_13_1330}
\|\hat{\bfx}_S - \bfx_S\|^2_2 - \|\hat{\bfx}_S - \bfy_S\|^2_2 \leqslant \|\bfx_{\bar{S}} - \bfy_{\bar{S}}\|^2_2 \Longleftrightarrow \|\hat{\bfx} - \bfx\|_2 \leqslant \|\hat{\bfx} - \bfy\|_2.
\end{equation}
Therefore,
\begin{equation*}
\underset{n \hspace*{2pt}\rightarrow \hspace*{2pt} \infty}{\text{lim}}  \mathbf{P}_{\bfx, \bfy}\left(\|\hat{\bfx} - \bfx \|_2 \leqslant \|\hat{\bfx} - \bfy\|_2\right)
= \underset{n \hspace*{2pt}\rightarrow \hspace*{2pt} \infty}{\text{lim}} \mathbf{P}_{\bfx, \bfy}\left(\|\hat{\bfx}_S - \bfx_S\|^2_2 - \|\hat{\bfx}_S - \bfy_S\|^2_2 \leqslant \|\bfx_{\bar{S}} - \bfy_{\bar{S}}\|^2_2\right) = 1.
\end{equation*}\\[5pt]

Note that $\bfx$ in the above derivation can be identified with the image of $\hat{\bfx}$ under the
conservative projection.
In order to prove the following equality (where $f := f_{con}$)
\begin{equation*}
\underset{n \hspace*{2pt}\rightarrow \hspace*{2pt} \infty}{\text{lim}}  \mathbf{P}_{\bfx, \bfy} (\|\hat{\bfx} - f(\hat{\bfx})\|_2 \leqslant \|\hat{\bfx} - \bfy\|_2) = 1,
\end{equation*}
it suffices to show that for all pairs $(\bfx, \bfy) \in D \times D$ the following implication holds true
\begin{equation}
\|\hat{\bfx} - \bfx \|_2 \leqslant \|\hat{\bfx} - \bfy\|_2 \Longrightarrow \|\hat{\bfx} - f(\hat{\bfx})\|_2 \leqslant \|\hat{\bfx} - \bfy\|_2.
\end{equation}
Namely, it holds:
\begin{equation*}
\|\hat{\bfx} - f(\hat{\bfx})\|^2 = {\underbrace{\|\hat{\bfx}_S - f_S(\hat{\bfx})\|^2}_{\leqslant \|\hat{\bfx}_{S} - \bfx_{S}\|^2}} + \underbrace{\|\hat{\bfx}_{\bar{S}} - f_{\bar{S}}(\hat{\bfx})\|^2}_{= 0} \overset{(*)}{\leqslant} \|\hat{\bfx}_S - \bfx_S\|^2 + \underbrace{\|\hat{\bfx}_{\bar{S}} - \bfx_{\bar{S}}\|^2}_{= 0} = \|\hat{\bfx} - \bfx\|^2 \leqslant \|\hat{\bfx} - \bfy\|^2,
\end{equation*}
where in step $(*)$ we used our Definition \ref{def_cons} of the conservative projection.
\end{proof}

\section{Proof of Theorem \ref{theorem_2}}
\begin{lemma}
\label{L_noise_diff}
Let $\bfx, \bfy, \hat{\bfx}, \bfepsilon \in \mathbb{R}^n$, where $\hat{\bfx} := \bfx + \bfepsilon$.
Then the following holds true:
\begin{equation}
\|\bfepsilon\| \leqslant \frac{1}{2}\|\bfx - \bfy\| \Longrightarrow \|\hat{\bfx} - \bfx \| \leqslant \|\hat{\bfx} - \bfy\|.
\end{equation}
\end{lemma}
\begin{proof}
It holds:
\begin{equation*}
\begin{aligned}
\|\hat{\bfx} - \bfx \| \leqslant \|\hat{\bfx} - \bfy\|
\Leftrightarrow \| \bfx +\bfepsilon - \bfx\| \leqslant \|\bfx + \bfepsilon - \bfy\|
\Leftrightarrow& \|\bfepsilon\| \leqslant \underbrace{\|(\bfx - \bfy) + \bfepsilon\|}_{\geqslant \|\bfx - \bfy \| - \|\bfepsilon\|} \\
\Leftarrow& \hspace*{3pt} \|\bfepsilon\| \leqslant \|\bfx - \bfy \| - \|\bfepsilon\|\\
\Leftrightarrow& \hspace*{3pt} \|\bfepsilon\| \leqslant \frac{1}{2}\|\bfx - \bfy \|\\
\end{aligned}
\end{equation*}
\end{proof}\\[5pt]

\noindent We now prove Theorem \ref{theorem_2}.
First we show:
\begin{equation}
\underset{n \hspace*{2pt}\rightarrow \hspace*{2pt} \infty}{\text{lim}} \mathbf{P}\left(\| \epsilon\| \leqslant \frac{1}{2}\| \bfx - \bfy \| \right) = 1.
\end{equation}
It holds:
\begin{equation*}
\begin{aligned}
& \hspace*{5pt} \| \epsilon\|^2 \leqslant \frac{1}{4}\| \bfx - \bfy \|^2\\
\Leftrightarrow& \hspace*{3pt} \underbrace{\left( \frac{1}{n}\|\bfepsilon\|^2 - (\mu_{\bfepsilon}^2 + \sigma^2_{\bfepsilon}) \right)}_{=: X} \leqslant \underbrace{\frac{1}{4} \left( \frac{1}{n}\|\bfx - \bfy\|^2 - 2\sigma^2 \right)}_{=: Y} + \underbrace{\left( \frac{1}{2}\sigma^2 - (\mu_{\bfepsilon}^2 + \sigma^2_{\bfepsilon})\right)}_{=: C}
\end{aligned}
\end{equation*}
It follows:
\begin{equation*}
\begin{aligned}
& \hspace*{3pt} \{X > Y + C\}\\
\subseteq& \hspace*{3pt} \{X > Y\} \\
=& \hspace*{3pt} \{X > Y, X \geqslant 0, Y \geqslant 0\} \cup \{X > Y, X \geqslant 0, Y < 0\} \cup \{X > Y, X < 0, Y < 0\} \\
&\cup \hspace*{3pt} \{X > Y, X < 0, Y \geqslant 0\}\\
=& \hspace*{3pt} \{|X| > |Y|, X \geqslant 0, Y \geqslant 0\} \cup \{|X| \geqslant 0, X \geqslant 0, Y < 0\} \cup \{|X| < |Y|, X < 0, Y < 0\}\\
=& \hspace*{3pt} \{|X| > |Y|, X \geqslant 0, Y \geqslant 0\} \cup \{|X| > 0, X > 0, Y < 0\} \cup \{X = 0, |Y| > 0\} \\
&\cup \hspace*{3pt} \{|X| < |Y|, X < 0, Y < 0\}\\
\subseteq& \hspace*{3pt} \{|X| > |Y|\} \cup \{|X| > 0\}  \cup \{|Y| > 0\}  \cup \{|X| < |Y|\}.
\end{aligned}
\end{equation*}
That is,
\begin{equation}
\label{e_0702_1839}
\mathbf{P}( X > Y + C ) \leqslant \mathbf{P}(\{|X| > |Y|\}) + \mathbf{P}(\{|X| > 0\}) + \mathbf{P}(\{|Y| > 0\})  + \mathbf{P}(\{|X| < |Y|\}).
\end{equation}
We can represent $\{|X| > |Y|\}$ and $\{|X| > 0\}$ as a union of disjoint sets according to
\begin{equation*}
\small
\begin{aligned}
\{|X| > |Y|\} = \left(\bigcup_{k \geqslant 1}\{k+1 \geqslant |Y| > k, |X| > k + 1\}\right) \cup \left(\bigcup_{k \geqslant 2}\{\frac{1}{k-1} \geqslant |Y| > \frac{1}{k}, |X| >  \frac{1}{k-1}\}\right)
\end{aligned}
\end{equation*}
and
\begin{equation*}
\small
\begin{aligned}
\{|X| > 0\} = \left(\bigcup_{k \geqslant 1}\{k+1 \geqslant |X| > k\}\right) \cup \left(\bigcup_{k \geqslant 2}\{\frac{1}{k-1} \geqslant |X| > \frac{1}{k}\}\right).
\end{aligned}
\end{equation*}
We get similar representation for $\{|X| < |Y|\}$ and $\{|Y| > 0\}$.
Now we consider the above derivations in the limit $n \rightarrow \infty$.
We look at the individual terms on the right-hand side of the equation in (\ref{e_0702_1839}) separately.
\begin{equation*}
\small
\begin{aligned}
& \hspace*{3pt} \underset{n \hspace*{2pt}\rightarrow \hspace*{2pt} \infty}{\text{lim}} \mathbf{P}( \{|X| > |Y|\} )\\
=& \hspace*{3pt}\underset{n \hspace*{2pt}\rightarrow \hspace*{2pt} \infty}{\text{lim}} \mathbf{P}(\left(\bigcup_{k \geqslant 1}\{k+1 \geqslant |Y| > k, |X| > k + 1\}\right) \cup \left(\bigcup_{k \geqslant 2}\{\frac{1}{k-1} \geqslant |Y| > \frac{1}{k}, |X| >  \frac{1}{k-1}\}\right))\\
\overset{(*)}{=}& \hspace*{3pt}\underset{n \hspace*{2pt}\rightarrow \hspace*{2pt} \infty}{\text{lim}} \underbrace{\sum_{k \geqslant 1} \mathbf{P}\left(k+1 \geqslant |Y| > k, |X| > k + 1\right)}_{\leqslant 1} + \underbrace{\sum_{k \geqslant 2} \mathbf{P}\left(\frac{1}{k-1} \geqslant |Y| > \frac{1}{k}, |X| >  \frac{1}{k-1}\right)}_{\leqslant 1}\\
\overset{(+)}{=}& \hspace*{3pt} \sum_{k \geqslant 1} \underset{n \hspace*{2pt}\rightarrow \hspace*{2pt} \infty}{\text{lim}}  \mathbf{P}\left(k+1 \geqslant |Y| > k, |X| > k + 1\right) + \sum_{k \geqslant 2} \underset{n \hspace*{2pt}\rightarrow \hspace*{2pt} \infty}{\text{lim}}  \mathbf{P}\left(\frac{1}{k-1} \geqslant |Y| > \frac{1}{k}, |X| >  \frac{1}{k-1}\right)\\
\leqslant& \hspace*{3pt} \sum_{k \geqslant 1} \underset{n \hspace*{2pt}\rightarrow \hspace*{2pt} \infty}{\text{lim}}  \mathbf{P}\left(|X| > k + 1\right) + \sum_{k \geqslant 2} \underset{n \hspace*{2pt}\rightarrow \hspace*{2pt} \infty}{\text{lim}}  \mathbf{P}\left(|Y| > \frac{1}{k}\right)\\
=& \hspace*{3pt} \sum_{k \geqslant 2} \underbrace{\underset{n \hspace*{2pt}\rightarrow \hspace*{2pt} \infty}{\text{lim}} \mathbf{P}(\left|\frac{1}{n}\|\bfepsilon\|^2 - (\mu_{\bfepsilon}^2 + \sigma^2_{\bfepsilon})\right| > k + 1)}_{= 0} + \sum_{k \geqslant 2} \underbrace{\underset{n \hspace*{2pt}\rightarrow \hspace*{2pt} \infty}{\text{lim}} \mathbf{P}(\left|\frac{1}{n}\|\bfx - \bfy\|^2 - 2\sigma^2\right| > \frac{1}{k})}_{= 0}\\
=& \hspace*{3pt} 0,
\end{aligned}
\end{equation*}
where in step $(*)$ we use the fact that all the sets are disjoint.
In step $(+)$ we swap the limit and the sum signs because the two sums are finite for all $n \in \mathbb{N}$.
In the last step we use the weak law of big numbers and ints variant for dependent variables with vanishing covariance.
Similarly, we get
\begin{equation}
\underset{n \hspace*{2pt}\rightarrow \hspace*{2pt} \infty}{\text{lim}} \mathbf{P}( |X| < |Y| ) = \underset{n \hspace*{2pt}\rightarrow \hspace*{2pt} \infty}{\text{lim}} \mathbf{P}( |X| > 0 ) = \underset{n \hspace*{2pt}\rightarrow \hspace*{2pt} \infty}{\text{lim}} \mathbf{P}( |Y| > 0 ) = 0.
\end{equation}
Therefore,
\begin{equation} 
\underset{n \hspace*{2pt}\rightarrow \hspace*{2pt} \infty}{\text{lim}} \mathbf{P}\left( \|\bfepsilon\| \leqslant \frac{1}{2}\|\bfx - \bfy\| \right) = \underset{n \hspace*{2pt}\rightarrow \hspace*{2pt} \infty}{\text{lim}} \mathbf{P}\left( X \leqslant Y + C \right) = 1 -  \underset{n \hspace*{2pt}\rightarrow \hspace*{2pt} \infty}{\text{lim}}  \mathbf{P}\left( X > Y + C \right) = 1.
\end{equation}
Applying Lemma \ref{L_noise_diff} to the above equation completes the proof:
\begin{equation} 
\underset{n \hspace*{2pt}\rightarrow \hspace*{2pt} \infty}{\text{lim}} \mathbf{P}\left(\| \hat{\bfx} - \bfx \| \hspace*{2pt}\leqslant \| \hat{\bfx} - \bfy \| \right) = 1.
\end{equation}

\section{Proof of Proposition \ref{prop_14031216}}
In order to prove Proposition \ref{prop_14031216} in the main body of the paper, we first introduce a set of auxiliary statements.
\begin{lemma}
\label{L_08041207}
\begin{itshape}
Let $f \colon \mathbb{R}^n \supseteq \mathcal{U} \longrightarrow \mathbb{R}^n$, $n \in \mathbb{R}$ be idempotent.
Then the equality
\begin{equation}
f(\bfx) = \bfx
\end{equation}
holds for all $\bfx \in f(\mathcal{U})$.
\end{itshape}
\end{lemma}
\begin{proof}
For any $\bfx \in f(\mathcal{U})$ there is $\bfz \in \mathcal{U}$ with $f(\bfz) = \bfx$.
It follows:
\begin{equation*}
f(\bfz) = \bfx \hspace*{6pt} \Longrightarrow \hspace*{6pt} f(f(\bfz)) = f(\bfx) \hspace*{6pt} \Longrightarrow \hspace*{6pt} f(\bfz) = f(\bfx) \hspace*{6pt} \Longrightarrow \hspace*{6pt} \bfx = f(\bfx),
\end{equation*}
where in the penultimate step we used the idempotency assumption
and in the last step $\bfx = f(\bfz)$.
\end{proof}


\begin{lemma}
\label{L_29021853}
\begin{itshape}
Let $\mathcal{D} \subseteq \mathbb{R}^n, n \in \mathbb{N}$ be a differentiable manifold and
$f, g \colon \mathbb{R}^n \rightarrow \mathbb{R}^n$ differentiable mappings such that $f(\bfx) = g(\bfx)$ for all $\bfx \in \mathcal{D}$.
Then the equality
\begin{equation}
D_{\bfx}f(\bfv) =  D_{\bfx}g(\bfv)
\end{equation}
holds true for all $\bfx \in \mathcal{D}$, $\bfv \in T_{\bfx}\mathcal{D}$.
\end{itshape}
\end{lemma}
\begin{proof}
Let $\bfx \in \mathcal{D}$ and $\bfv \in T_{\bfx}\mathcal{D} \setminus \{\bf0\}$.
Since $\mathcal{D}$ is a differentiable manifold there exists a parametrization $\phi \colon \mathcal{U} \rightarrow \mathbb{R}^m$, $\bfx \in \mathcal{U}$ such that
$f \circ \phi^{-1}, g \circ \phi^{-1}$ are differentiable. Because $D_{\bfq}\phi^{-1}(\mathbb{R}^m) = T_{\bfx}\mathcal{D}$ for $\bfq = \phi(\bfx)$ holds,
there exists $\bfw \in \mathbb{R}^m$ such that $D_{\bfq}\phi^{-1}(\bfw) = \bfv$.
It follows:
\begin{equation*}
\begin{aligned}
& \hspace*{6pt} f(\bfx) = g(\bfx) \hspace*{10pt}\text{ for all } \bfx \in \mathcal{D}\\
\Longrightarrow& \hspace*{6pt}f \circ \phi^{-1}(\bfq) = g \circ \phi^{-1}(\bfq) \hspace*{10pt}\text{ for all } \bfq \in \phi(\mathcal{U})\\
\Longrightarrow& \hspace*{6pt}D_{\bfq}f \circ \phi^{-1} =  D_{\bfq}g \circ \phi^{-1}\\
\Longrightarrow& \hspace*{6pt}D_{\phi^{-1}(\bfq)}f \circ D_{\bfq}\phi^{-1} =  D_{\phi^{-1}(\bfq)}g \circ D_{\bfq}\phi^{-1}\\
\Longrightarrow& \hspace*{6pt}D_{\bfx}f \circ D_{\bfq}\phi^{-1}(\bfw) =  D_{\bfx}g \circ D_{\bfq}\phi^{-1}(\bfw) \hspace*{10pt}\text{ for all } \bfw \in \mathbb{R}^m\\
\Longrightarrow& \hspace*{6pt}D_{\bfx}f(\bfv) =  D_{\bfx}g(\bfv).
\end{aligned}
\end{equation*}
\end{proof}

\begin{lemma}
\label{L_04030930}
\begin{itshape}
Let $\mathcal{D} \subseteq \mathbb{R}^n$ be a differentiable manifold. Consider an orthogonal projection $f$ and another projection $g$ onto $\mathcal{D}$.
The following holds true for all $\bfx \in \mathcal{D}$:
\begin{equation}
\|D_{\bfx}f\|_2 \hspace*{3pt} \leqslant \hspace*{1pt} \|D_{\bfx}g\|_2.
\end{equation}
\end{itshape}
\end{lemma}
\begin{proof}
Let $\bfx \in \mathcal{D}$ be fixed.
Because $\mathbb{R}^n = T_{\bfx}\mathcal{D} \oplus (T_{\bfx}\mathcal{D})^{\perp}$,
each $\bfv \in \mathbb{R}^n$ can be represented as a sum of two vectors $\bfv = \bfu + \bfw$, where $\bfu \in T_{\bfx}\mathcal{D}$, $\bfw \in (T_{\bfx}\mathcal{D})^{\perp}$.
In particular, it holds $\bfu^{\top}\bfw = 0$.
It follows:
\begin{equation*}
\begin{aligned}
\|D_{\bfx}f\|_2 &\hspace*{2pt}= \sup_{\bfv \neq 0}\frac{\|D_{\bfx}f(\bfv)\|_2}{\|\bfv\|_2} =  \sup_{\bfu + \bfw \neq 0}\frac{\|D_{\bfx}f(\bfu + \bfw)\|_2}{\|\bfu + \bfw\|_2} \hspace*{3pt} \overset{(*)}{=}  \sup_{\bfu + \bfw \neq 0}\frac{\|D_{\bfx}g(\bfu)\|_2}{\|\bfu + \bfw\|_2}
\overset{(**)}{=}  \sup_{\bfu + \bfw \neq 0, \bfu \neq 0}\frac{\|D_{\bfx}g(\bfu)\|_2}{\|\bfu + \bfw\|_2} \\
&\overset{(***)}{\leqslant}  \sup_{\bfu + \bfw \neq 0, \bfu \neq 0}\frac{\|D_{\bfx}g(\bfu)\|_2}{\|\bfu\|_2} =  \sup_{\bfu \neq 0}\frac{\|D_{\bfx}g(\bfu)\|_2}{\|\bfu\|_2} = \|D_{\bfx}g\|_2,
\end{aligned}
\end{equation*}
where in step $(*)$ we used the orthogonality of $f$, according to which $D_{\bfx}f(\bfw) = 0$ for all $\bfw \in (T_{\bfx}\mathcal{D})^{\perp}$, and $D_{\bfx}f(\bfu) =  D_{\bfx}g(\bfu)$ for all $\bfu \in T_{\bfx}\mathcal{D}$ (according to Lemma \ref{L_29021853}).
In step $(**)$, we used the fact that the supremum of a corresponding term is achieved for $\bfu \neq \bf0$, since $\|D_{\bfx}g(\bf0)\|_2 = 0$ and we can always find a $\bfu \neq \bf0$ with $\|D_{\bfx}g(\bfu)\|_2, \|\bfu + \bfw\|_2 > 0$.
In step $(***)$ we used the following argument: $\bfu^{\top}\bfw = 0 \Longrightarrow \|\bfu + \bfw\|_2 \hspace*{2pt}\geqslant \|\bfu\|_2$.
\end{proof}

\begin{corollary}
\label{cor_08040943}
\begin{itshape}
Let $\mathcal{D}$ be a differentiable manifold. Consider an orthogonal projection $f$ and another projection $g$ onto $\mathcal{D}$.
The following holds true for all $\bfx \in \mathcal{D}$:
\begin{equation}
\|D_{\bfx}f\|_F \hspace*{3pt} \leqslant \hspace*{1pt}\|D_{\bfx}g\|_F.
\end{equation}
\end{itshape}
\end{corollary}
\begin{proof}
The proof follows directly from Lemma \ref{L_04030930}.
Namely, based in the geometric interpretation, the singular values of a matrix $A$ correspond to the length
of the major axis of the ellipsoid $E_A$, which is given by the image of the euclidian unit-ball under the linear transformation $A$.
If the inequality $\|A\bfx\| \leqslant \|B\bfx\|$ holds for all $\bfx$, this implies that $E_A$ is completely contained within $E_B$.
Let us denote by $\lambda_1, ..., \lambda_r$ and $\mu_1, ..., \mu_k$, $r \leqslant k$ the positive singular values (in descending order) of $A$ and $B$, respectively.
Then this implies $\lambda_i \leqslant \mu_i$ for all $i \in \{1, ..., r\}$.
In particular, it follows:
\begin{equation*}
\|A\|_F = \sqrt{\lambda_1^2 + ... \lambda_n^2 } \leqslant \sqrt{\mu_1^2 + ... \mu_n^2 } = \|B\|_F.
\end{equation*}
In particular, since $D_{\bfx}f(\bfv) = D_{\bfx}g(\bfv)$ for all $\bfv \in T_{\bfx}\mathcal{D}$, the singular vectors lying in $T_{\bfx}\mathcal{D}$ (and the corresponding singular values)
will be the same for both matrices.
\end{proof}\\[5pt]

\noindent Now we can prove the statement in Proposition \ref{prop_14031216}.
Note that any idempotent mapping $f \in C^1(\mathcal{U})$ satisfies $\|f(\bfx) - \bfx\| = 0$ and
(according to Lemma \ref{L_29021853} and Corollary \ref{cor_08040943}) $\|D_{\bfx}f^*\|_2 \hspace*{3pt} \leqslant \hspace*{1pt} \|D_{\bfx}f\|_2$ and
$\|D_{\bfx}f^*\|_F \hspace*{3pt} \leqslant \hspace*{1pt}\|D_{\bfx}f\|_F$ for all $\bfx \in \mathcal{D}$ and $f \in C^1(\mathcal{U})$.
Since $f^*$ minimizes both terms of the sum in (\ref{E_08040946}), it is an optimal solution.

\section{Proof of Proposition \ref{prop_14031813}}
Consider the plot in Figure \ref{fig_bauer11} illustrating the unit circles with respect to $l^p$-norm for $p \in \{1, 2, \infty\}$ centered around the point $\hat{\bfx} \in \mathbb{R}^2$.
Now choose some $p$ and increase the radius of the unit circle around $\hat{\bfx}$ until it touches the set $D$,
which here represents the submanifold of normal examples.
Note that the touching points $\bfy^*_p$ correspond to orthogonal projections of $\hat{\bfx}$ onto $D$ with respect to the $l^p$-norm.
For $p = 1$ there are two possible cases for the position of touching points independently of the shape of $D$.
In the first case, the $l^1$-unit circle touches $D$ at one of its corners.
Without loss of generality assume that this corner is the point $(r, 0)$ for some $r > 0$.
This implies that $\hat{x}_1 \neq y^*_1$ and $\hat{x}_2 = y^*_2$, that is, $S_{1} = \{1\}$.
However, since all norms are equal at the corners of $l^1$-circle, it also holds $1 \in S_{p}$ for $p \geqslant 2$.
On the other hand, if the touching point lies on a side (or on an edge in the higher-dimensional case)
of the $l_1$-circle, it implies $S_{1} = \{1, 2\}$.
Obviously, we also get $S_{\infty} = \{1, 2\}$.
Consider now an orthogonal projection of $D$ onto $\mathbb{R}^2$ spanned by a pair of axes .
Note that each $l^p$-circle for $2 \leqslant p < \infty$ at such a touching point has a non-zero curvature towards the interior of the circle.
See Figure \ref{fig_bauer11} for an illustration.
This implies that the touching point with respect to the $l^p$-norm must lie within
the rectangular triangle shaped by the touching points $\bfy^*_{1}$, $\bfy^*_{\infty}$, and the lines through these points parallel to the two axes.
In particular, for all $p < q$ the touching point $\bfy^*_q$
lies in a smaller triangle shaped by the points $\bfy^*_{p}$, $\bfy^*_{\infty}$ and the lines through these points parallel to the coordinated axes.
For the plot in Figure \ref{fig_bauer11}, this means that $\bfy^*_q$
will lie on a red curve between the points $\bfy^*_{p}$ and $\bfy^*_{\infty}$
implying $S_{q} = \{1, 2\}$.
Since this argument is independent of our choice of the axes, it proves the statement $(a)$.

\begin{figure}
\centering
\includegraphics[scale = 0.6]{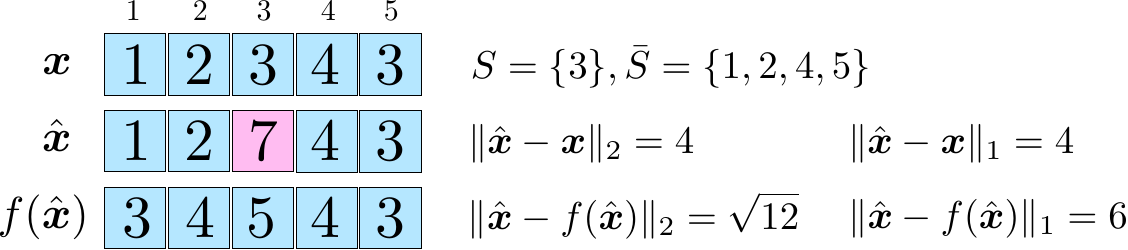}
\caption{
Illustration of a counterexample for the claim that orthogonal projections maximally preserve normal regions in the inputs.
Here, $\hat{\bfx} \in \mathbb{Z}^5$ is the modified version of the original input $\bfx \in \mathcal{D}$ according to the partition $S, \bar{S}$
and $f(\hat{\bfx})$ denotes the orthogonal projection of $\hat{\bfx}$ onto $\mathcal{D}$ with respect to the $l_2$-norm.
This example also shows that orthogonality property is dependent on our choice of the distance metric.
}
\label{fig_bauer17}
\end{figure}

In order to prove the statement in $(b)$, we provide a counterexample for the equality.
Consider an example of a Markov chain describing a simple random walk on the one-dimensional grid of integers
illustrated in Figure \ref{fig_bauer17}.
Precisely, we consider a sequence of i.i.d variables $(\xi_i)_{i \in \mathbb{N}}$ with values in $\{-1, 1\}$
and define a Markov chain $(x_n)_{n \in \mathbb{N}}$ according to $x_n := \sum_{i = 1}^n \xi_i$ with $x_0 := 0$.
Finally, we define the set of normal examples $\mathcal{D} \subseteq \{(x_{n+1}, ..., x_{n+5}) \in \mathbb{Z}^5 \hspace*{3pt}|\hspace*{3pt} n \in \mathbb{N}\}$
as the set of all feasible configurations for subsequences of length $5$.
We can see that there is a mismatch between $S(\hat{\bfx}, \bfx)$ and $S(\hat{\bfx}, f(\hat{\bfx}))$.
This provides a counterexample for the claim
that orthogonal projection maximally preserves uncorrupted regions.


\section{Proof of Proposition \ref{P_09041241}}
\begin{figure}[t]
\centering
\includegraphics[scale = 0.5]{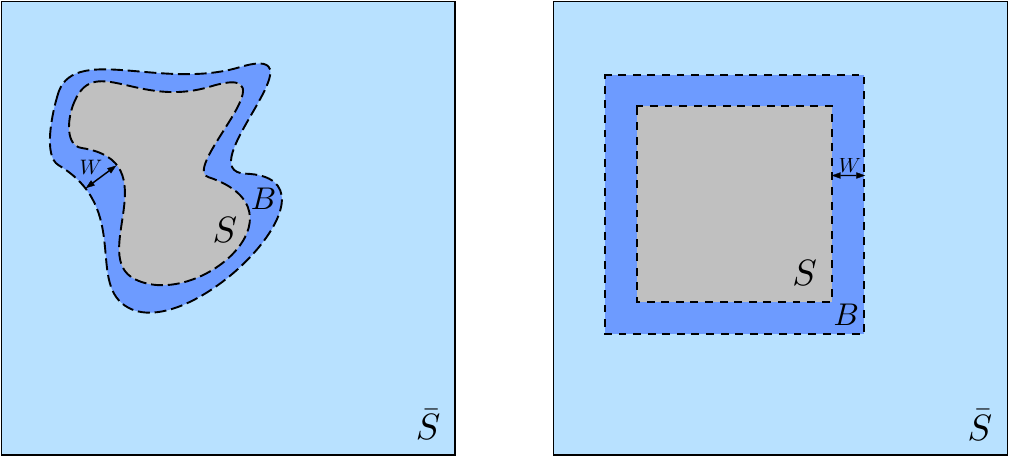}
\caption{
Illustration of the concept of a transition set on two examples with different shapes. Each of the two images represent an MRF $\bfx = (x_1, ..., x_n)$, $n \in \mathbb{N}$
of the Markov order $K \in \mathbb{N}$
with nodes corresponding to the individual pixels with values from a finite set of states $x_i \in I$, $|I| < \infty$.
The grey area marks the corrupted region $S \subseteq \{1, ..., n\}$,
where the union of the dark blue and light blue areas is the complement $\bar{S} := \{1, ..., n\} \setminus S$ marking the normal region.
The dark blue part of $\bar{S}$ corresponds to the transition set $B \subseteq \bar{S}$.
$W \leqslant |I|^K$ denotes (loosely speaking) the width at the thickest part of the tube $B$ around $S$.
}
\label{fig_bauer13}
\end{figure}
In the following, we naturally generalize some properties of Markov chains to the two-dimensional case of MRFs,
where the nodes are organized in a grid-like structure similarly to the Ising model.
Here, we omit a formal definition and provide only an intuition, which is sufficient for our purposes.
Informally, we introduce the notion of the order of an MRF
by relating the definition for Markov chains to the width
of the corresponding Markov blanket.
Furthermore, we generalize the notion of irreducible Markov chains to irreducible MRFs by assuming that each state
is reachable from any other state on the chain-subgraphs of the MRF, where each node $x_j$ on the chain
has a higher distance (with respect to the topology of the MRF graph G) to the start node $x_1$
than all of the previous nodes $x_i$ according to the metric $d_G(\cdot, \cdot)$.
That is, $i < j \Longrightarrow d_G(x_i, x_1) < d_G(x_j, x_1)$.
Note that the notion of a state at position $i$ on the path now also involves the values
of the variables in the neighborhood (corresponding to the Markov blanket) of $x_i$ in the MRF.
Essentially, this property is covered by our assumption on the variables to be identically distributed with vanishing covariance.
Based on this informal extension we introduce an auxiliary concept we refer to as the \emph{transition set}.
Consider first a Markov chain of the order $K \in \mathbb{N}$ with a finite set of possible states $x_i \in I, |I| < \infty$.
Then the maximal number of steps required to reach a state $j \in I$ from a state $i \in I$ is upper-bounded by $|I|^K$,
that is, by a constant independent of the graph size of the MRF.
Namely, when traveling from one node to another we see at most $|I|^K$ combinations of states $(x_{n-K}, ..., x_{n-1}, x_n)$ with $\mathbf{P}(x_n | x_{n-K}, ..., x_{n-1}) > 0$,
which locally affect our path. The shortest path would have no redundant configurations $(x_{n-K}, ..., x_{n-1})$.
Therefore, given two patterns $(x_{m+1}, ..., x_{m+r-1})$ and $(x_{m+r}, ..., x_{m+n})$,
we can always find (for sufficiently large numbers $r, n \in \mathbb{N}$) a sequence $B = (x_{m+r-b}, ..., x_{m+r-1})$, $b \leqslant |I|^K$
such that $\mathbf{P}(x_{m+1}, ..., x_{m+r-b+1}, x_{m+r-b}, ..., x_{m+r-1}, x_{m+r}, ..., x_{m+n}) > 0$.
This example can be generalized to the MRF by considering transitions between the individual nodes,
which respect the values of the surrounding variables in their neighborhoods (or Markov blankets).
Basically, this corresponds to considering Markov chains with an increased set of states upper-bounded by $|I|^{K^2}$.
We refer to the set of pixels $B$ as a \emph{transition set}.
See Figure \ref{fig_bauer13} for an illustration.\\[5pt]

\noindent Now we can prove the statement of Proposition \ref{P_09041241}.
In the following, we use the notation $\bfy^* := F(\hat{\bfx})$.
Consider now $\bfy := ({\bfy^*_S}^{\top}, \bfy_B^{\top}, \bfx_{\bar{S} \setminus B}^{\top})^{\top}$
for some feasible $\bfy_B$.
Provided $\min \{|S|, |\bar{S}| \}$ is sufficiently greater than $K$, we can always find a feasible transition set $B$
such that $\bfy \in B$.
It follows:
\begin{equation*}
\begin{aligned}
 \|\hat{\bfx} - \bfy^* \|_p \leqslant \| \hat{\bfx} - \bfy\|_p
\overset{\ref{l1}}{\Longleftrightarrow}& \hspace*{3pt} \|\hat{\bfx}_S - \bfy^*_S\|^p_p + \|\hat{\bfx}_{\bar{S}} - \bfy^*_{\bar{S}}\|^p_p \leqslant \|\hat{\bfx}_S - \bfy^*_S\|^p_p + \|\hat{\bfx}_B - \bfy_B\|^p_p + \underbrace{\|\hat{\bfx}_{\bar{S} \setminus B} - \bfx_{\bar{S} \setminus B}\|^p_p}_{= 0} \\
 \Longleftrightarrow& \hspace*{3pt} \|\hat{\bfx}_{\bar{S}} - \bfy^*_{\bar{S}}\|^p_p \leqslant \|\hat{\bfx}_B - \bfy_B\|^p_p \leqslant |B|\\
 \Longrightarrow& \hspace*{3pt} \|\bfx_{\bar{S}} - F_{\bar{S}}(\hat{\bfx})\|^p_p  \leqslant |B|,
\end{aligned}
\end{equation*}
where in the penultimate step we used $x_i, y_i \in [0, 1]$.

Now we show $|B| \in O(\sqrt{|S|})$.
Without loss of generality, we assume that $S$ and $B$ have a square-shape with equal-length sides as illustrated in the right part of Figure \ref{fig_bauer13}.
That is, each side of the grey square has the length $\sqrt{|S|}$.
The area of the region corresponding to $B$ is then given by $|B| = 4 \sqrt{|S|} W + 4 W^2$, where $W$ is a graph independent constant.
Furthermore, by putting the last equation in the inequality $|B| < |S|$ and solving the resulting quadratic equation,
we get a criterion $\min \{|S|, |\bar{S}| \} > (2 + \sqrt{8})^2 W^2$, which guarantees the existence of a feasible transition set $B$.

\section{Visualization of Qualitative Improvement in Reconstruction when using Model II over I}
\begin{figure}
\centering
\includegraphics[scale = 0.85]{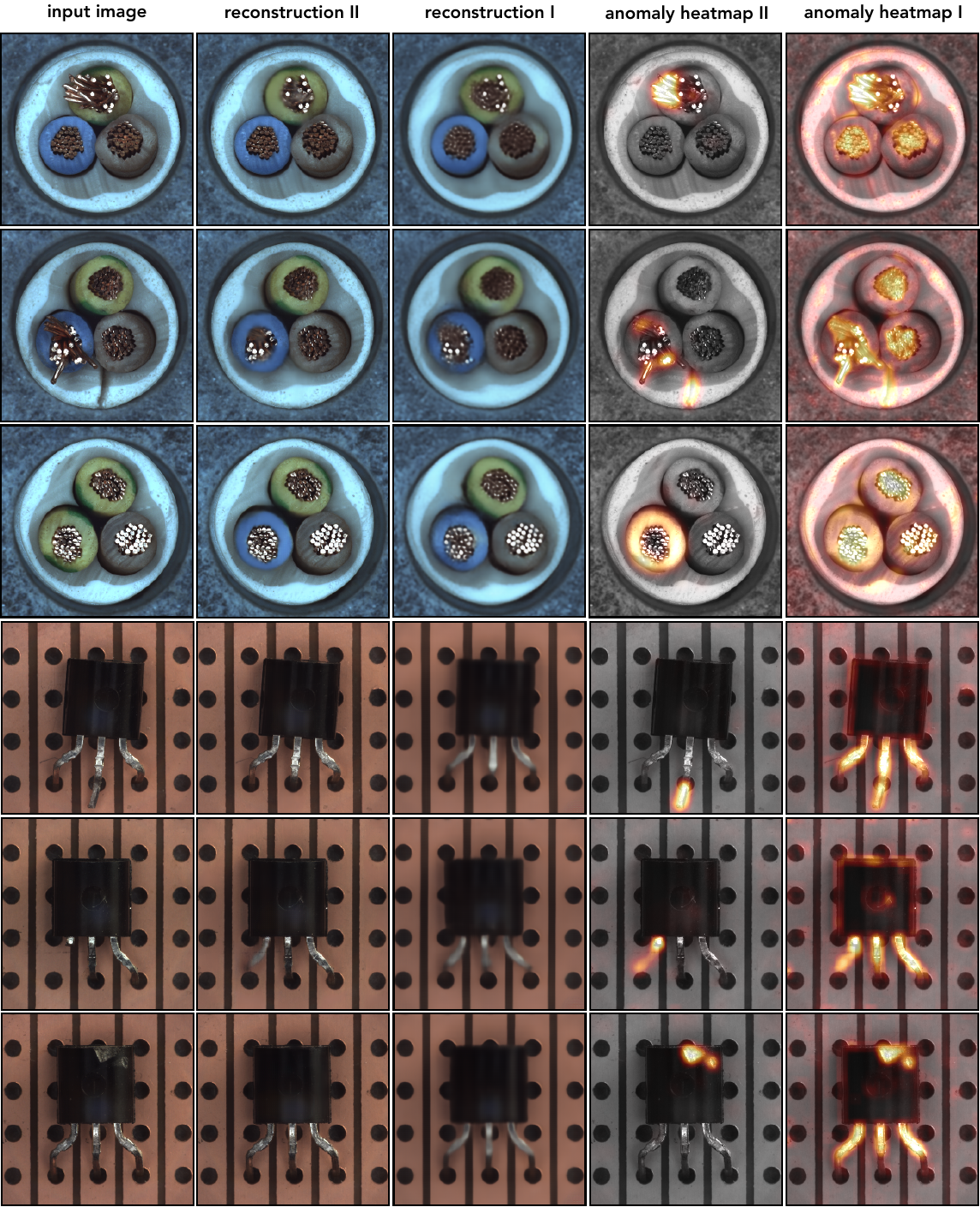}
\caption{
Illustration of the qualitative improvement when using Model II over I.
We show six examples: three from the category "cable" and three from the category "transistor" of the MVTec AD dataset.
The images in each row represent the original image, the reconstruction produced by Model II, the reconstruction produced by Model I,
the anomaly heatmap by Model II and the anomaly heatmap by Model I, respectively.
Note the dramatic improvement in the quality of heatmaps.
}
\label{fig_bauer14}
\end{figure}
\begin{figure}
\centering
\includegraphics[scale = 0.85]{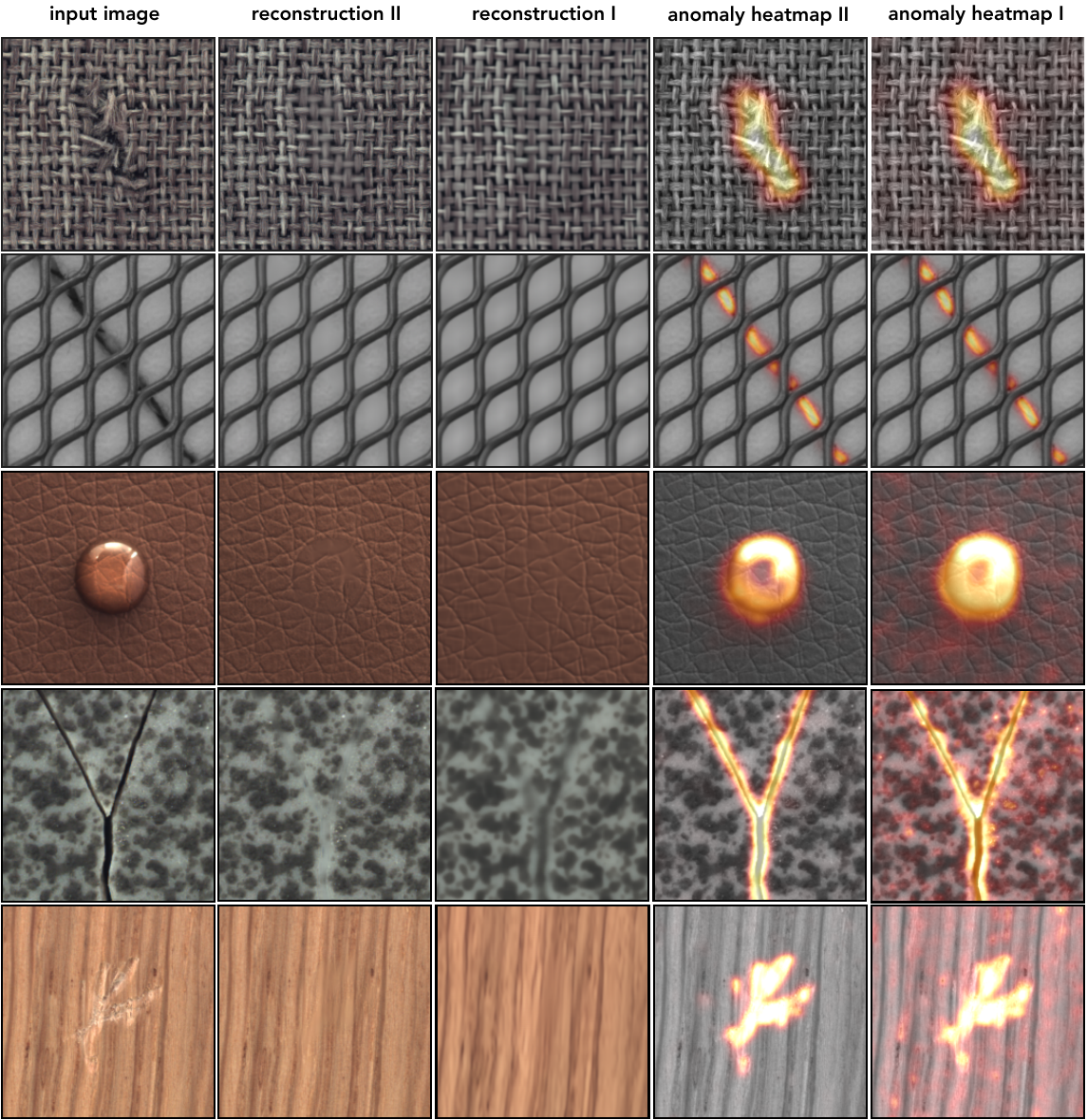}
\caption{
Illustration of the qualitative improvement of the reconstruction when using Model II over I on texture categories of the MVTec AD dataset.
We show five examples: one from each category "carpet", "grid", "leather", "tile" and "wood".
The images in each row represent the original image, the reconstruction produced by Model II, the reconstruction produced by Model I,
the anomaly heatmap by Model II and the anomaly heatmap by Model I, respectively.
Despite the reconstruction improvement, the resulting heatmaps show similar quality.
}
\label{fig_bauer15}
\end{figure}
As outlined in the main body of the paper,
in order to achieve high detection and localization performance based on the reconstruction error,
the model must replace corrupted regions in the input images with a different content.
At the same time, it is important to reproduce uncorrupted regions as accurate as possible
to reduce the chance of false positive detections.
In this sense, the overall reconstruction quality of the original content during training
correlates strongly with the ability of the corresponding model to detect anomalous samples.
We compared the reconstruction performance of two model architecture, which we refer to as Model I and Model II in the paper.
During our empirical evaluation we observed that using Model I,
while on average achieving good reconstruction, can sometimes result in higher number of false positive detections than Model II.
For example, it struggles to reproduce the normal regions, which are characterized by frequent change of the gradient along the neighboring pixel values,
on two object categories "cable" and "transistor" from the MVTec AD dataset.
In contrast, Model II accurately reconstructs these regions resulting in a significant reduction of false positive detections.
See Figure \ref{fig_bauer14} for further examples.
Similarly, we observed an improvement in reconstruction quality for texture categories,
The corresponding heatmaps, however, show similar quality. 
We provide a few examples in Figure \ref{fig_bauer15}.

\end{document}